\documentclass[11pt]{article}

\usepackage[final]{acl}
\usepackage{booktabs}
\usepackage{multirow}
\usepackage{threeparttable}

\usepackage{times}
\usepackage{latexsym}
\usepackage{amsmath}
\usepackage{amssymb}
\usepackage[T1]{fontenc}
\usepackage[utf8]{inputenc}

\usepackage{microtype}

\usepackage{inconsolata}
\usepackage{tabularx} %表格自动换行
\usepackage[table]{xcolor}
\usepackage[dvipsnames,table]{xcolor}
\usepackage{bm}

\usepackage{graphicx}

\newtheorem{assumption}{Assumption}
\title{IDEA: An Interpretable and Editable Decision-Making Framework\\
for LLMs via Verbal-to-Numeric Calibration}

\author{
  Yanji He$^{1}$,
  Yuxin Jiang$^{2}$,
  Yiwen Wu$^{1}$,
  Bo Huang$^{1}$,
  Jiaheng Wei\thanks{Corresponding author.}$^{1}$,
  Wei Wang\footnotemark[1]$^{1,3}$ \\
  $^{1}$The Hong Kong University of Science and Technology (Guangzhou) \\
  $^{2}$Huawei Technologies Co.,Ltd,
  $^{3}$The Hong Kong University of Science and Technology \\
  \texttt{\{yhe720, ywu240\}@connect.hkust-gz.edu.cn, jiang.yuxin2@huawei.com,} \\
  \texttt{bhuangas@connect.ust.hk, jiahengwei@hkust-gz.edu.cn, weiwcs@ust.hk}
}

\begin{document}
\maketitle
\begin{abstract}
Large Language Models are increasingly deployed for decision-making, yet their adoption in high-stakes domains remains limited by miscalibrated probabilities, unfaithful explanations, and inability to incorporate expert knowledge precisely. We propose \textbf{IDEA}, a framework that extracts LLM decision knowledge into an interpretable parametric model over semantically meaningful factors. Through joint learning of verbal-to-numerical mappings and decision parameters via EM, correlated sampling that preserves factor dependencies, and direct parameter editing with mathematical guarantees, IDEA produces calibrated probabilities while enabling quantitative human-AI collaboration. Experiments across five datasets show IDEA with Qwen-3-32B (78.6\%) outperforms DeepSeek R1 (68.1\%) and GPT-5.2 (77.9\%), achieving perfect factor exclusion and exact calibration—precision unattainable through prompting alone. The implementation is publicly available.\footnote{Source code is available at \url{https://github.com/leonbig/IDEA}.}
\end{abstract}

\section{Introduction}

\begin{figure}[t]
    \centering
    \includegraphics[width=0.49\textwidth]{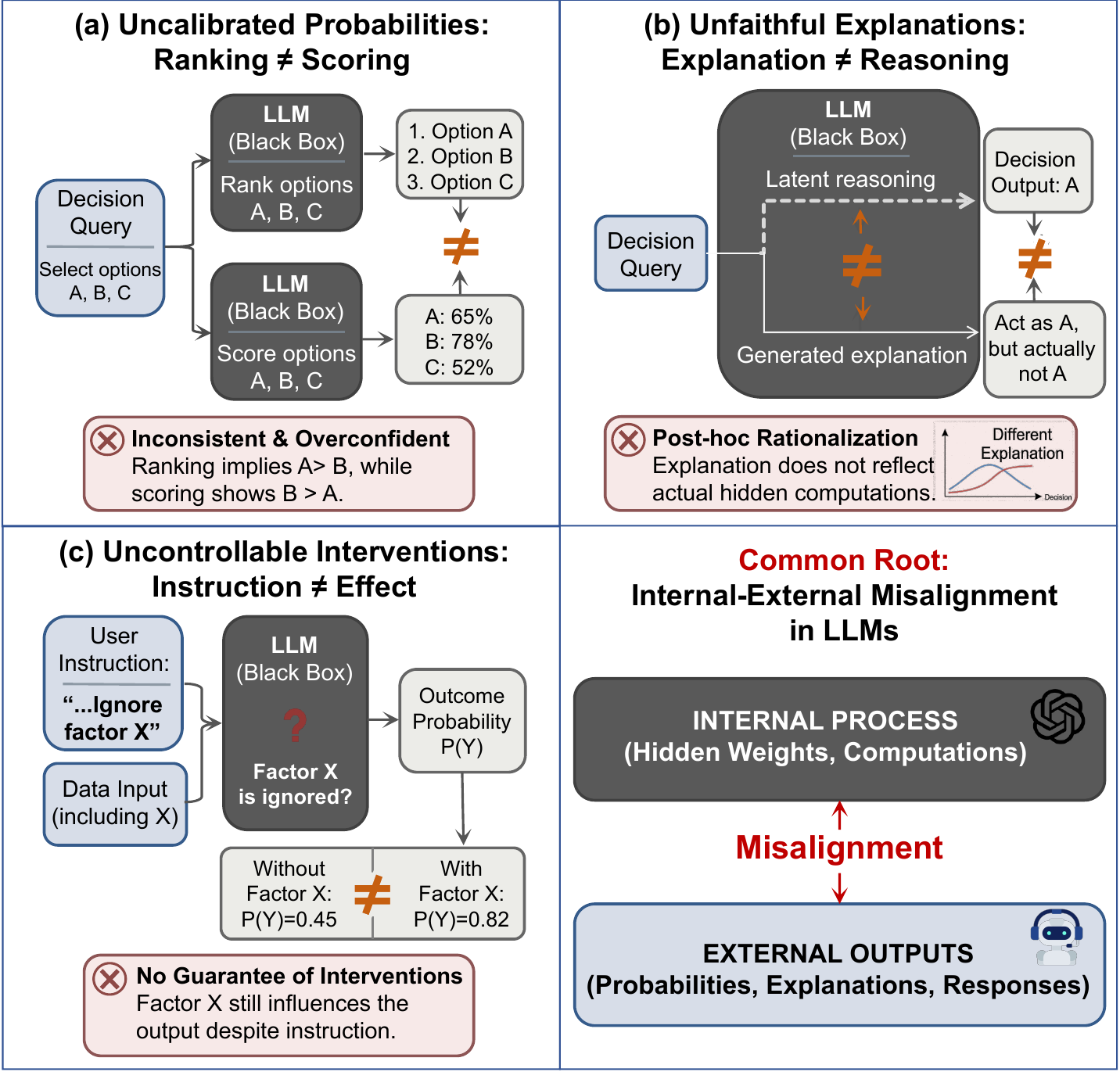}
    \caption{Three manifestations of the trust deficit in LLM-based decision making, stemming from a common root: internal-external misalignment. (a) \textbf{Uncalibrated probabilities}: ranking and scoring yield inconsistent orderings for the same options. (b) \textbf{Unfaithful explanations}: generated rationales diverge from the latent reasoning that actually determines outputs. (c) \textbf{Uncontrollable interventions}: natural language instructions fail to guarantee behavioral compliance—factor X continues to influence predictions despite explicit exclusion instructions.}
    \label{fig:trust_deficit}
\end{figure}

Large Language Models (LLMs) are increasingly deployed for automated decision making~\citep{DBLP:journals/corr/abs-2303-04129}, yet their use in high-stakes domains like financial investment remains limited by a fundamental trust deficit: stakeholders cannot reliably verify, audit, or intervene in the decision process.
This deficit stems from three challenges. First, real-world decisions demand calibrated probabilities, yet LLMs produce overconfident and inaccurate estimates under uncertainty~\citep{DBLP:conf/iclr/XiongHLLFHH24}. Second, stakeholders require faithful explanations, but generated rationales often serve as post-hoc rationalizations rather than reflecting actual reasoning~\citep{DBLP:conf/nips/TurpinMPB23, lanham2023measuringfaithfulnesschainofthoughtreasoning}. Third, current systems lack quantitative frameworks to integrate domain expert insights beyond simple prompting (Figure~\ref{fig:trust_deficit}).

Existing approaches fail to resolve these issues. \textbf{Logit-based} methods extract probabilities from output distributions~\citep{kadavath2022languagemodelsmostlyknow, tian2023just}, but conflate next-token confidence with decision uncertainty and remain black-box transformations. \textbf{Self-explanation} approaches like chain-of-thought prompting~\citep{DBLP:conf/nips/Wei0SBIXCLZ22} and self-consistency decoding \citep{DBLP:conf/iclr/0002WSLCNCZ23} produce readable justifications but demonstrably lack faithfulness. \textbf{Factor-based} frameworks decompose decisions through intermediate variables: DeLLMa ~\citep{liu2024dellma} uses decision-theoretic structures with direct numerical elicitation, and BIRD~\citep{DBLP:conf/iclr/0013ZLR25} elicits verbal probabilities over factors. However, DeLLMa relies on LLMs to produce precise numerical utilities (precisely the capability we identify as unreliable) and BIRD assumes factor independence with fixed verbal-to-numerical mappings from prior literature, losing both calibration accuracy (our w/o EM ablation shows $-6.8\%$ avg F1) and natural factor correlations. Neither framework supports quantitative parameter editing with mathematical guarantees. These limitations reflect a deeper misalignment between LLMs' internal computations and their external outputs.

We pursue a different path grounded in two observations (validated in Section~\ref{sec:probing}): (i) while LLMs cannot reliably produce precise numerical probabilities, they can generate decision-relevant factors from their broad knowledge; and (ii) LLMs exhibit greater consistency when producing verbal probability expressions (e.g., "likely," "unlikely") than exact numeric estimates—a consequence of training on human text where such phrases are abundant while precise probabilities are rare.

Rather than making the internal inference process of LLMs transparent, we aim to extract their knowledge into a form that is inherently transparent. Specifically, we propose \textbf{IDEA} (An \textbf{I}nterpretable and E\textbf{d}itable D\textbf{e}cision-M\textbf{a}king Framework) by introducing an intermediate representation: a set of semantically meaningful decision factors $\mathcal{F} = \{F_1, \ldots, F_N\}$. Given a decision query $Q$ comprising a scenario $S$ (e.g., loan approval) and specific conditions $C$ (e.g., applicant information), our goal is to estimate $\mathbb{P}(O_i \mid Q)$ for each possible outcome $O_i$. Rather than eliciting this probability directly, we assume these factors are jointly sufficient to determine the decision outcome. Under this assumption, the target probability admits a decomposition:
\begin{equation}
    \mathbb{P}(O_i \mid Q) = \sum_{\mathbf{f} \in \mathcal{F}^*} \mathbb{P}(O_i \mid \mathbf{f}) \cdot \mathbb{P}(\mathbf{f} \mid Q),
    \label{eq:intro_factorization}
\end{equation}
where $\mathcal{F}^*$ denotes the space of all factor value assignments (factor configurations), and $\mathbf{f}$ denotes a specific assignment. This decomposition separates two components: the \emph{decision model} $\mathbb{P}(O_i \mid \mathbf{f})$, which maps factor configurations to outcome probabilities, and the \emph{factor inference} $\mathbb{P}(\mathbf{f} \mid Q)=\mathbb{P}(\mathbf{f} \mid C)$, which determine the factor values from the entire query, that is, from the specific condition. The key insight is that the decision model $\mathbb{P}(O_i \mid \mathbf{f})$ operates over a low-dimensional, semantically structured space. This enables learning an interpretable model whose parameters directly quantify each factor's contribution, allowing inspection, verification, and modification of the decision process.

Two technical challenges arise that distinguish IDEA from prior factor-based approaches. First, verbal probability expressions (e.g., "very unlikely," "likely") are inherently ambiguous with unknown numerical mappings, yet learning the decision model requires numerical labels. Unlike BIRD~\citep{DBLP:conf/iclr/0013ZLR25}, which fixes mappings from psychological literature, we resolve this through joint estimation of the verbal-to-numerical mapping and model parameters via EM, treating unknown probabilities as latent variables. This addresses a fundamental circularity: learning the decision model requires numerical labels, but determining what numerical values verbal expressions represent requires the decision model. Second, inference under partial information requires marginalizing over uncertain factors while preserving their dependencies. Unlike methods assuming conditional independence~\citep{DBLP:conf/iclr/0013ZLR25}, we sample joint configurations directly from the LLM, maintaining natural correlations (e.g., high income with stable employment).

The resulting framework offers three properties that directly address the trust deficit:
\begin{itemize}
    \item \textbf{Calibrated Probability Estimation.} The framework produces well-calibrated probabilities through joint learning of verbal-to-numerical mappings and decision model parameters.
    \item \textbf{Semantic Interpretability.} Each parameter directly quantifies a specific factor's contribution, enabling domain experts to inspect and contest the decision logic.
    \item \textbf{Quantitative Human-AI Collaboration.} Users can edit parameters to incorporate domain knowledge or enforce constraints with mathematically precise effects—unattainable through prompting alone.
\end{itemize}

\section{Related Work}

\paragraph{Probability Elicitation and Calibration.}
Prior work extracts probabilities via token logits ~\citep{kadavath2022languagemodelsmostlyknow} or verbalized confidence \citep{tian2023just}, but both exhibit poor calibration \citep{DBLP:conf/kdd/LiuCDC0025}. BIRD ~\citep{DBLP:conf/iclr/0013ZLR25} elicits verbal probabilities over factors but assumes independence and fixes verbal mappings. Bayesian approaches model uncertainty through posterior distributions ~\citep{gal2016dropout, DBLP:conf/iclr/KuhnGF23}, but remain computationally prohibitive for LLMs. Ensemble methods like self-consistency ~\citep{DBLP:conf/iclr/0002WSLCNCZ23} improve calibration but treat decisions as black boxes. We jointly learn calibrated mappings while preserving factor dependencies and enabling parameter inspection.

\paragraph{Interpretable Models and Explanations.}
Concept Bottleneck Models (CBMs) route predictions through concept layers ~\citep{koh2020concept}, with variants addressing concept uncertainty ~\citep{DBLP:conf/icml/KimJPKY23}, annotation costs ~\citep{DBLP:conf/iclr/OikarinenDNW23, DBLP:conf/cvpr/YangPZJCY23}, and LLM integration ~\citep{DBLP:conf/iclr/SunOUW25}. However, these require task-specific training or assume concept independence. Post-hoc methods like SHAP ~\citep{DBLP:conf/nips/LundbergL17} explain predictions but cannot modify behavior. Our framework provides CBM-style interpretability while uniquely enabling direct parameter editing with mathematical guarantees.

\section{Framework Overview}

\begin{figure*}[t]
    \centering
    \includegraphics[width=\textwidth]{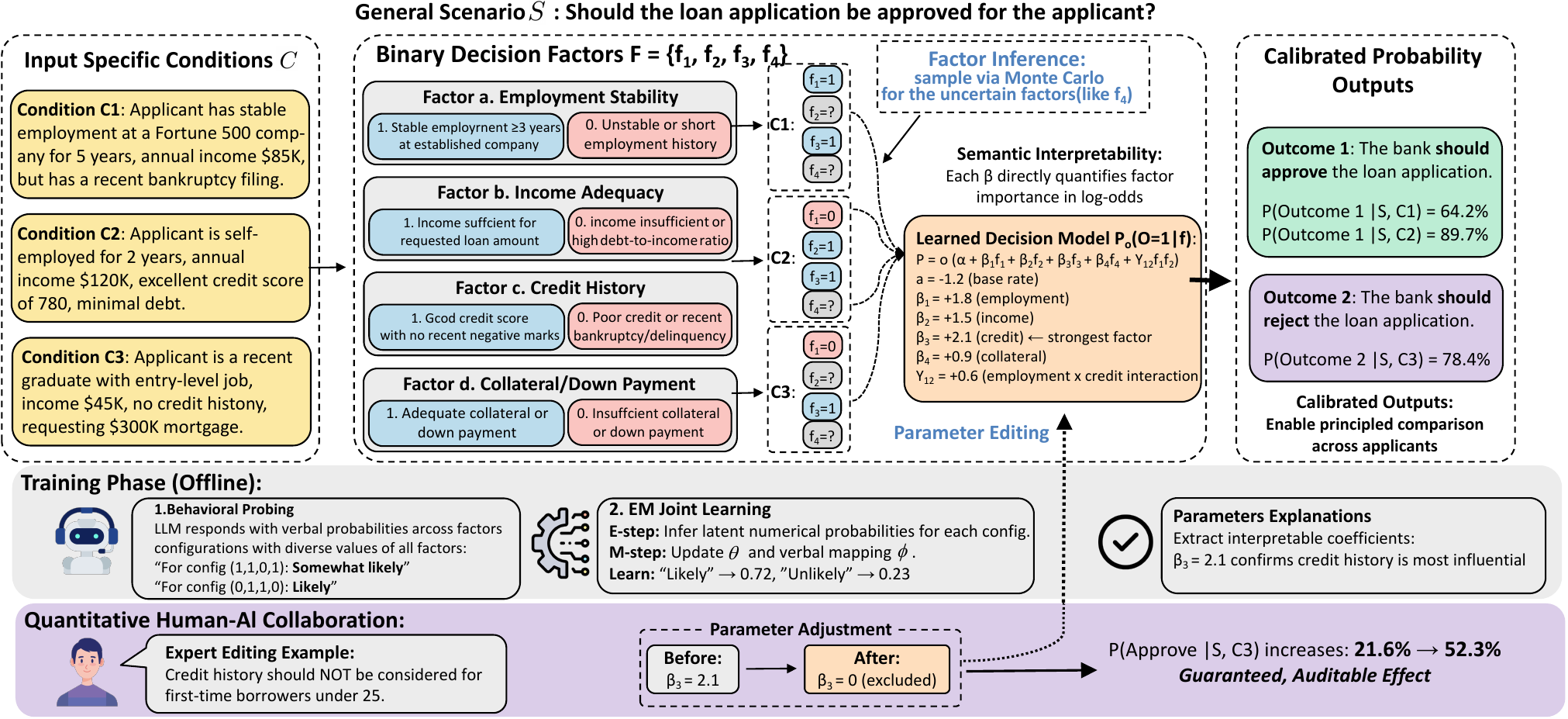}
    \caption{The IDEA framework illustrated on a loan approval task. Given applicant conditions $C$, the framework extracts binary factor values, handles uncertainty via Monte Carlo sampling, and produces calibrated probabilities through a learned decision model $P_{\boldsymbol{\theta}}(O = 1 \mid \mathbf{f})$ with interpretable coefficients. The offline phase (bottom left) jointly learns model parameters and verbal-to-numerical mappings via EM. Expert intervention (bottom right) enables precise, auditable parameter edits---e.g., excluding credit history by setting $\beta_3 = 0$ shifts approval probability from 21.6\% to 52.3\%.}
    \label{fig:framework}
\end{figure*}

We now formalize the problem and present our framework (illustrated in Figure~\ref{fig:framework}). The general formulation in Equation~\eqref{eq:intro_factorization} accommodates arbitrary outcome spaces and factor domains. In this work, we focus on \textbf{binary decisions} with \textbf{binary factors}: the outcome space is $\mathcal{O} = \{0, 1\}$, and each factor $F_j \in \{0, 1\}$ indicates whether that factor's value is positive. Our goal is to estimate $\mathbb{P}(O=1 \mid Q)$ for a query $Q$ comprising a scenario $S$ and a specific condition $C$. 

\subsection{Technical Objectives and Assumptions}

Under the binary setting, the factor configuration space becomes $\mathcal{F}^* = \{0,1\}^N$. Our objectives are: 
\begin{itemize}
    \item \textbf{Learn the decision model} $\mathbb{P}(O_i \mid \mathbf{f})$ as an interpretable parametric model $P_{\boldsymbol{\theta}}(O_i \mid \mathbf{f})$ from LLM behavioral data.
    \item \textbf{Perform factor inference} $\mathbb{P}(\mathbf{f} \mid C)$ that determine factor values from specific conditions.
\end{itemize}

Finally, compute the final probability via Equation~\eqref{eq:intro_factorization}. Our framework relies on some key assumptions: 
\begin{assumption}[Factor Completeness]\label{ass:fc}
% \paragraph{Factor Completeness.} 
The identified factors $\mathcal{F}$ are jointly sufficient to determine the decision outcome: $\mathbb{P}(O_i \mid C, S) = \mathbb{P}(O_i \mid \mathbf{f})$, where $\mathbf{f}$ is the factor configuration determining the values of factors by specific condition $C$. 
\end{assumption}
Assumption \ref{ass:fc} ensures no decision-relevant information is lost in the factor representation.
\begin{assumption}[Verbal Probability Consistency]\label{ass:vpc}
The LLM's verbal probability responses reflect an underlying numerical probability through a consistent, monotonic mapping. 
\end{assumption}
While the exact mapping is unknown, Assumption \ref{ass:vpc} assumes it can be learned from behavioral data. We empirically validate this monotonic-consistency requirement in Appendix~\ref{app:ordinal}, finding 87.6\% ordinal agreement across 7680 configuration pairs.

\subsection{Framework Pipeline}

As depicted in Figure \ref{fig:framework}, our framework extracts LLM decision knowledge into an interpretable parametric model over a semantically meaningful factor space. The methodology comprises five components organized into two phases.
\paragraph{Training Phase (Offline).} Given a decision query $Q = (S, C)$: 1) \textit{Factor Identification}: Construct binary factors set $\mathcal{F} = \{F_1, \dots, F_N\}$ and verify their discriminability and coverage to satisfy Assumption~\ref{ass:fc}. 2) \textit{Behavioral Probing}: Query the LLM across plenty of factor configurations to collect verbal probability responses, yielding dataset $\mathcal{D} = \{(\mathbf{f}^{(k)}, V^{(k)})\}_{k=1}^K$. 3) \textit{Joint Estimation}: Apply Expectation-Maximization to simultaneously learn the decision model $P_{\bm{\theta}}(O \mid \mathbf{f})$ and the verbal-to-numerical mapping $\phi : \mathcal{V} \rightarrow [0, 1]$ on the dataset $\mathcal{D}$.

\paragraph{Inference Phase (Online).} 1) \textit{Factor Determination}: Extract observable factor values from conditions $C$, partitioning factors into observed $(\mathcal{O})$ and uncertain $(\mathcal{U})$ subsets. 2) \textit{Marginalization}: Sample coherent completions for uncertain factors to preserve dependencies, then compute the final probability through Monte Carlo marginalization over the learned decision model.

\paragraph{Expert Intervention.} The framework additionally supports quantitative human-AI collaboration through direct parameter manipulation, enabling precise editing of relative factor importance with mathematically guaranteed effects.

\section{Methodology} 

\subsection{Factor Identification and Verification}
\label{sec:factor_id}

For a decision scenario $S$ with outcomes $\mathcal{O} = \{O_1, O_2\}$, we construct a binary factor set $\mathcal{F} = \{F_1, \ldots, F_N\}$ through statement-driven elicitation rather than direct generating. To encourage comprehensive exploration of the decision space, we prompt the LLM to generate 20 comprehensive situational descriptions that would cause an outcome to occur. From these generated statements, we summarize candidate factors and formulated as binary values where $F_j = 1$ and $F_j = 0$ encodes whether a semantically distinct aspect is positive or negative. The factor set undergoes iterative verification by LLMs against two criteria: (i) \emph{discriminability}, each factor's values must differentially support distinct outcomes, and (ii) \emph{coverage}, the factors must jointly subsume all decision-relevant information in the specific condition $C$. Factors failing discriminability are reformulated or discarded; unmapped information units in $C$ trigger factor expansion. This loop iterates until convergence, yielding a factor set satisfying Assumption~\ref{ass:fc}.

\subsection{Behavioral Probing}
\label{sec:probing}

With a verified factor set $\mathcal{F}$, we now collect behavioral data that reveals how the LLM maps factor configurations to outcome likelihoods. This dataset forms the empirical foundation for learning the decision model in Section~\ref{sec:em_training}. We construct the training dataset by systematically querying the LLM across the factor configuration space $\mathcal{F}^* = \{0,1\}^N$. For tractable factor spaces ($2^N \leq 256$), we probe exhaustively. For larger spaces, we employ uniform random sampling of 256 configurations, which provides sufficient coverage for estimating main effects and salient interactions.

\paragraph{Verbal Probability Elicitation.} For each configuration $\mathbf{f}^{(k)}$, we construct a hypothetical scenario and query the LLM to assess outcome likelihood. Following~\citep{DBLP:conf/iclr/0013ZLR25}, we adopt a seven-level ordinal scale: $\mathcal{V} = $\{\textit{very unlikely}, \textit{unlikely}, \textit{somewhat unlikely}, \textit{neutral}, \textit{somewhat likely}, \textit{likely}, \textit{very likely}\}, ordered from $v_1$ to $v_7$ by increasing likelihood. This process yields a behavioral dataset $\mathcal{D} = \{(\mathbf{f}^{(k)}, V^{(k)})\}_{k=1}^K$, where $V^{(k)} \in \mathcal{V}$ is the LLM's verbal response to configuration $\mathbf{f}^{(k)}$.

\subsection{Joint Estimation via EM Algorithm}
\label{sec:em_training}

The behavioral probing procedure yields a dataset $\mathcal{D} = \{(\mathbf{f}^{(k)}, V^{(k)})\}_{k=1}^K$, where each observation pairs a factor configuration $\mathbf{f}^{(k)} \in \{0,1\}^N$ with the LLM's verbal probability response $V^{(k)} \in \mathcal{V}$. Our goal is to learn two components simultaneously:
\begin{itemize}
    \item A \textbf{decision model} $P_{\boldsymbol{\theta}}(O=1 \mid \mathbf{f})$ that maps factor configurations to numerical probabilities, parameterized by $\boldsymbol{\theta}$.
    \item A \textbf{verbal-to-numerical mapping} $\phi: \mathcal{V} \rightarrow [0,1]$ that converts verbal expressions to calibrated probabilities.
\end{itemize}

This presents a "chicken-and-egg problem": if the mapping $\phi$ were known, we could directly regress model predictions against numerical targets $\phi(V^{(k)})$, and vice versa.

We therefore employ the Expectation-Maximization (EM) algorithm, which iteratively computes the posterior distribution of latent probabilities (E-step) and updates parameters to maximize the expected complete-data likelihood (M-step). EM alternation guarantees monotonic improvement in the marginal likelihood and convergence to a local optimum.

\subsubsection{Model Specification}

We instantiate the decision model as a logistic regression with main effects and pairwise interactions. Recall that a factor configuration $\mathbf{f} = (f_1, \dots, f_N) \in \{0, 1\}^N$ assigns a binary value to each factor, where $f_j \in \{0, 1\}$ denotes the value of factor $F_j$. The decision model takes the form:
\begin{equation}\resizebox{1.05\hsize}{!}{$
P_{\boldsymbol{\theta}}(O = 1 \mid \mathbf{f}) = \sigma\left(\alpha + \sum_{j=1}^N \beta_j f_j + \sum_{1 \leq i < j \leq N} \gamma_{ij} f_i f_j\right)
$},
\label{eq:proxy_model}
\end{equation}
where $\sigma(\cdot)$ is the sigmoid function, $\alpha$ is the intercept, $\beta_j$ quantifies the main effect of factor $F_j$, and $\gamma_{ij}$ captures the interaction between factors $F_i$ and $F_j$. The full parameter vector is $\bm{\theta} = (\alpha, \bm{\beta}, \bm{\gamma})$ with $\bm{\beta} = (\beta_1, \dots, \beta_N)$ and $\bm{\gamma} = \{\gamma_{ij}\}_{i<j}$.

This functional form balances expressiveness with interpretability: main effects $\{\beta_j\}$ directly quantify each factor's contribution in log-odds, while interactions $\{\gamma_{ij}\}$ capture synergies or conflicts between factor pairs. To promote sparse, interpretable interaction structure, we apply elastic net regularization exclusively to the interaction coefficients: $\Omega(\boldsymbol{\gamma}) = \lambda_1 \|\boldsymbol{\gamma}\|_1 + \lambda_2 \|\boldsymbol{\gamma}\|_2^2$.
\label{eq:elastic_net}
The $\ell_1$ penalty induces sparsity, driving negligible interactions to exactly zero; the $\ell_2$ penalty ensures numerical stability when interactions are correlated. Main effects $\{\beta_j\}$ and intercept $\alpha$ remain unregularized to preserve their direct interpretability.

\subsubsection{Estimation Algorithm}
We formulate the estimation problem by introducing latent variables representing the unknown numerical probabilities. For each observation $(\mathbf{f}^{(k)}, V^{(k)})$, let $p^{(k)} \in [0, 1]$ denote the latent probability that the LLM's verbal response $V^{(k)}$ implicitly represents. These latent variables bridge verbal expressions and numerical targets, enabling joint estimation via the Expectation-Maximization (EM) algorithm \citep{dempster1977maximum}.

\paragraph{E-step.} Given current estimates $(\boldsymbol{\theta}^{(t)}, \phi^{(t)})$, we compute the posterior distribution of each latent probability $p^{(k)}$. Two sources of information constrain this distribution: the model prediction $P_{\boldsymbol{\theta}}(O=1 \mid \mathbf{f}^{(k)})$ and the verbal indication $\phi(V^{(k)})$. Under a Gaussian likelihood model with precision parameters $\tau_{\theta}$ and $\tau_{\phi}$, the posterior mean takes the form of a precision-weighted combination:
\begin{equation}
    \mathbb{E}[p^{(k)} \mid \mathbf{f}^{(k)}, V^{(k)}] = \frac{\tau_{\theta} \cdot P_{\boldsymbol{\theta}}(\mathbf{f}^{(k)}) + \tau_{\phi} \cdot \phi(V^{(k)})}{\tau_{\theta} + \tau_{\phi}}.
\end{equation}

\paragraph{M-step.} We update both components to maximize the expected complete-data log-likelihood. The model parameters $\boldsymbol{\theta}$ are updated by minimizing a composite objective:
\begin{equation}\resizebox{1.00\hsize}{!}{$
    \mathcal{L}(\boldsymbol{\theta}) = \sum_{k} \left( P_{\boldsymbol{\theta}}(\mathbf{f}^{(k)}) - \mathbb{E}[p^{(k)}] \right)^2 + \lambda \cdot \mathcal{L}_{\text{rank}}(\boldsymbol{\theta}) + \Omega(\boldsymbol{\gamma}),
    $}
\end{equation}
where $\mathcal{L}_{\text{rank}}$ is a directional hinge consistency loss ~\citep{DBLP:conf/icml/BurgesSRLDHH05} that penalizes violations of the ordinal structure (i.e., configurations eliciting "likely" should yield higher probabilities than those eliciting "unlikely"). The verbal mapping $\phi$ is updated by minimizing the squared deviation from posterior means, subject to monotonicity constraints that preserve semantic ordering ~\citep{wallsten1986measuring}.
We initialize $\phi$ with canonical mappings from psychological literature ~\citep{Budescu2009Improving}. Iteration terminates when the Q-function (expected complete-data log-likelihood) change falls below $10^{-4}$. Full derivations appear in Appendix~\ref{appendix:em_details}.

\subsection{Inference Under Uncertainty}
\label{sec:inference}
Given a new query $Q=(S,C)$, inference follows the decomposition in Equation~\eqref{eq:intro_factorization} through three steps.

\paragraph{Factor Determination.} For each factor $F_j$, we query the LLM to decide whether specific condition $C$ implies: $F_j=1$, $F_j=0$, or does not determine $F_j$. This partitions factors into observed factors $\mathcal{O}$ (with determined values $\mathbf{f}_{\mathcal{O}}$) and uncertain factors $\mathcal{U}$. A human evaluation on 150 instances confirms this partition is reliable (91.5\% accuracy on the observed-vs-uncertain distinction for Qwen-3-32B; see Appendix~\ref{app:factor-det}).

\paragraph{Joint Sampling.} When $|\mathcal{U}| > 0$, we must estimate the joint distribution over uncertain factors. Unlike independence assumptions in prior work ~\citep{DBLP:conf/iclr/0013ZLR25}, we preserves natural factor correlations (e.g., high income correlating with stable employment) by sampling $T$ completions from the LLM conditioned on $\mathbf{f}_{\mathcal{O}}$ and $C$. High temperature ($\tau \geq 1.0$) ensures sample diversity.

\paragraph{Marginalization.} The final probability is computed via Monte Carlo marginalization:
\begin{equation}
P(O=1\mid Q)\approx \frac{1}{T}\sum_{t=1}^T
P_{\boldsymbol{\theta}}\!\left(O=1\mid \mathbf{f}_{\mathcal{O}},\mathbf{f}^{(t)}_{\mathcal{U}}\right).
\label{eq:mc_estimation}
\end{equation}
This estimator is unbiased, with standard error $O(1/\sqrt{T})$. In our experiments, we use $T=50$ samples, which provides stable estimates while remaining computationally efficient.

\subsection{User Parameter Editing}
\label{sec:manipulation}
The decision model's transparent structure enables expert intervention impossible through prompting. We support two modes: structural edits (adding or removing factors) and quantitative edits to factor influence.
Users can directly add or remove factors during training. Because the model operates over the explicit factor space  $\mathcal{F}$ rather than opaque internal activations, such modifications are auditable guaranteed.
\paragraph{Average Marginal Effects.}

Experts often need to adjust the relative importance of factors quantitatively. However, logistic coefficients operate in log-odds space and do not directly represent probability changes. To express constraints in probability space, we employ the \textbf{Average Marginal Effect (AME)}---the expected probability change when a factor switches from 0 to 1, averaged over all configurations of other factors:
\begin{equation}
\text{AME}_k = \frac{1}{2^{N-1}} \sum_{\mathbf{x}_{-k} \in \{0,1\}^{N-1}} \Delta_k(\mathbf{x}_{-k}), \label{eq:ame}
\end{equation}
where $\Delta_k(\mathbf{x}_{-k}) = P_{\boldsymbol{\theta}}(O=1 \mid f_k=1, \mathbf{x}_{-k}) - P_{\boldsymbol{\theta}}(O=1 \mid f_k=0, \mathbf{x}_{-k})$ and $\mathbf{x}_{-k}$ denotes all factor assignments except $k$. AME values are directly interpretable as average percentage-point changes in outcome probability. AME is widely employed to interpret parameter importance in logistic regression, as it bridges the gap between abstract statistical coefficients and human cognitive intuition ~\citep{mood2010logistic}.

\paragraph{Constrained Adjustment.} 
To enforce expert constraints such as relative importance (e.g., $\text{AME}_2 = \rho \cdot \text{AME}_1$) while minimizing unintended side effects, we solve:
\begin{equation}
\begin{aligned}
\min_{\bm{\theta}'} \quad & \sum_{j \in \{1,\ldots,N\} \setminus \{1,2\}} \left(\text{AME}_j({\boldsymbol{\theta}}') - \text{AME}_j({\boldsymbol{\theta}}) \right)^2 \\
\text{s.t.} \quad & \mathbb{E}[z_{{\boldsymbol{\theta}}'}] = \mathbb{E}[z_{{\boldsymbol{\theta}}}], \\
& \text{AME}_2({\boldsymbol{\theta}}') = \rho \cdot \text{AME}_1({\boldsymbol{\theta}}'),
\end{aligned}
\label{eq:constrained_opt}
\end{equation}
where $z_{\boldsymbol{\theta}} = \alpha + \sum_j \beta_j f_j + \sum_{i<j} \gamma_{ij} f_i f_j$ is the logit under ${\boldsymbol{\theta}}$, and expectations are taken over $\{0, 1\}^N$ (or a domain-weighted distribution when base rates are known). Since $\text{AME}({\boldsymbol{\theta}})$ is differentiable with respect to ${\boldsymbol{\theta}}$, we solve this with Sequential Quadratic Programming. The resulting interventions are \emph{verifiable} (constraints can be checked exactly), \emph{predictable} (other AMEs are minimally perturbed), and \emph{reversible} (the original ${\boldsymbol{\theta}}$ can be restored).

\section{Experiments}
\subsection{Experimental Setup}
\paragraph{Datasets.}
We evaluate on five datasets across two categories. \textit{Complex Decisions}: BIGDATA22~\citep{soun2022accurate} (1,472 instances) predicts stock movements from market signals and tweets; Statlog German Credit ~\citep{hans_hofmann_2023} (1,000 instances) assesses loan applicant creditworthiness. \textit{Reasoning}: COMMON2SENSE~\citep{DBLP:conf/acl/SinghWHAWMP21} (3,672 instances) tests commonsense reasoning; PLASMA~\citep{DBLP:journals/corr/abs-2305-19472} (1,178 instances) evaluates procedural planning; TODAY ~\citep{DBLP:conf/acl/FengZ0JR23} (895 instances) assesses temporal reasoning. Following~\citep{DBLP:conf/iclr/0013ZLR25}, all instances are converted to binary decision queries.
\paragraph{Models and Baselines.}
We use Qwen-3-4B/8B/32B ~\citep{DBLP:journals/corr/abs-2505-09388}. Baselines include: (1) \textit{Ablations}—\textbf{w/o Inter} removes interaction terms ($\gamma_{ij} = 0$), testing whether factor interactions capture meaningful decision patterns; \textbf{w/o EM} uses fixed verbal-to-numerical mappings from prior literature ~\citep{Budescu2009Improving}, bypassing joint calibration; \textbf{w/o MC} deterministically assigns uncertain factors (randomly with $p = 0.5$ when uninformed), testing whether correlated sampling preserves important factor dependencies. (2) \textit{Existing methods}—\textbf{Logit} extracts probabilities from token logits; \textbf{Vanilla} ~\cite{DBLP:conf/iclr/0002WSLCNCZ23} directly elicits verbalized probabilities; \textbf{CoT} ~\citep{DBLP:conf/nips/Wei0SBIXCLZ22} applies chain-of-thought before verbalization; \textbf{PWC} performs pairwise comparison without explicit probabilities (Experiment II only, where its comparative framing provides structural advantage). Hyperparameters are in the Appendix \ref{appendix:hyperparams}.

\begin{table*}[htbp]
    \small
    \centering
    \footnotesize
    \begin{tabularx}{\textwidth}{@{}lXcccccc|ccccc@{}}
        \toprule
        \textbf{Model} & \textbf{Method} & \textbf{BIG} & \textbf{Crd} & \textbf{C2S} & \textbf{PLS} & \textbf{TOD} & \textbf{Avg} & \textbf{F1 $C_1$} & \textbf{F1 $C_2$} & \textbf{F1 Eq} & \textbf{Avg} \\
        \midrule
        N/A & Random & 50.0 & 50.0 & 50.0 & 50.0 & 50.0 & 50.0 & 0.333 & 0.333 & 0.333 & 0.333 \\
        DeepSeek R1 & CoT & 51.8 & 57.0 & 88.7 & 75.4 & 67.7 & 68.1 & 0.318 & 0.335 & 0.205 & 0.286 \\
        GPT-5.2 & CoT & 67.8 & 68.2 & 95.0 & 82.9 & 75.8 & 77.9 & 0.455 & 0.438 & 0.295 & 0.402 \\
        \midrule
        \multirow{8}{*}{Qwen-3-4B} 
        & CoT & 51.4 & \textbf{57.5} & \textbf{94.2} & 75.1 & 68.1 & 69.3 & 0.352 & 0.325 & 0.261 & 0.313 \\
        & Logit & 52.4 & 54.9 & 88.8 & 75.6 & 70.2 & 68.4 & 0.281 & 0.292 & 0.238 & 0.270 \\
        & Vanilla & 56.7 & 56.1 & 91.7 & 82.0 & 70.4 & 71.4 & 0.392 & 0.361 & 0.284 & 0.346 \\
        & BIRD & 55.1 & 53.1 & 88.0 & 75.3 & 71.4 & 68.6 & 0.398 & 0.415 & 0.190 & 0.334 \\
        & PWC & - & - & - & - & - & - & 0.542 & \textbf{0.610} & \textbf{0.415} & \textbf{0.522} \\
        & \cellcolor{SeaGreen!10}IDEA w/o Inter & \cellcolor{SeaGreen!10}54.4 & \cellcolor{SeaGreen!10}53.4 & \cellcolor{SeaGreen!10}93.0 & \cellcolor{SeaGreen!10}79.3 & \cellcolor{SeaGreen!10}68.6 & \cellcolor{SeaGreen!10}69.7 & \cellcolor{SeaGreen!10}0.522 & \cellcolor{SeaGreen!10}0.494 & \cellcolor{SeaGreen!10}0.358 & \cellcolor{SeaGreen!10}0.458 \\
        & \cellcolor{SeaGreen!10}IDEA w/o EM & \cellcolor{SeaGreen!10}55.2 & \cellcolor{SeaGreen!10}52.7 & \cellcolor{SeaGreen!10}91.8 & \cellcolor{SeaGreen!10}75.7 & \cellcolor{SeaGreen!10}68.1 & \cellcolor{SeaGreen!10}68.7 & \cellcolor{SeaGreen!10}0.505 & \cellcolor{SeaGreen!10}0.476 & \cellcolor{SeaGreen!10}0.168 & \cellcolor{SeaGreen!10}0.383 \\
        & \cellcolor{SeaGreen!10}IDEA w/o MC & \cellcolor{SeaGreen!10}54.3 & \cellcolor{SeaGreen!10}53.5 & \cellcolor{SeaGreen!10}88.7 & \cellcolor{SeaGreen!10}75.2 & \cellcolor{SeaGreen!10}68.5 & \cellcolor{SeaGreen!10}68.0 & \cellcolor{SeaGreen!10}0.536 & \cellcolor{SeaGreen!10}0.510 & \cellcolor{SeaGreen!10}0.332 & \cellcolor{SeaGreen!10}0.459 \\
        & \cellcolor{SeaGreen!20}IDEA & \cellcolor{SeaGreen!20}\textbf{56.8} & \cellcolor{SeaGreen!20}53.3 & \cellcolor{SeaGreen!20}93.1 & \cellcolor{SeaGreen!20}\textbf{83.9} & \cellcolor{SeaGreen!20}\textbf{70.7} & \cellcolor{SeaGreen!20}\textbf{71.6} & \cellcolor{SeaGreen!20}\textbf{0.575} & \cellcolor{SeaGreen!20}0.553 & \cellcolor{SeaGreen!20}0.385 & \cellcolor{SeaGreen!20}0.504 \\
        \midrule
        \multirow{8}{*}{Qwen-3-8B}
        & CoT & 50.7 & 58.5 & 93.9 & 75.9 & 66.8 & 69.2 & 0.365 & 0.352 & 0.248 & 0.322 \\
        & Logit & 56.1 & 58.5 & 91.3 & 76.4 & 70.2 & 70.5 & 0.322 & 0.314 & 0.236 & 0.291 \\
        & Vanilla & 59.1 & 58.6 & 91.9 & 78.1 & 70.7 & 71.7 & 0.418 & 0.392 & 0.301 & 0.370 \\
        & BIRD & 56.2 & 55.1 & 91.3 & 73.3 & 72.4 & 69.7 & 0.552 & 0.567 & 0.346 & 0.488 \\
        & PWC & - & - & - & - & - & - & 0.615 & 0.528 & 0.415 & 0.519 \\
        & \cellcolor{SeaGreen!10}IDEA w/o Inter & \cellcolor{SeaGreen!10}50.8 & \cellcolor{SeaGreen!10}58.9 & \cellcolor{SeaGreen!10}93.0 & \cellcolor{SeaGreen!10}80.2 & \cellcolor{SeaGreen!10}70.6 & \cellcolor{SeaGreen!10}70.7 & \cellcolor{SeaGreen!10}0.675 & \cellcolor{SeaGreen!10}0.660 & \cellcolor{SeaGreen!10}0.501 & \cellcolor{SeaGreen!10}0.612 \\
        & \cellcolor{SeaGreen!10}IDEA w/o EM & \cellcolor{SeaGreen!10}52.8 & \cellcolor{SeaGreen!10}60.2 & \cellcolor{SeaGreen!10}89.4 & \cellcolor{SeaGreen!10}76.6 & \cellcolor{SeaGreen!10}68.3 & \cellcolor{SeaGreen!10}69.5 & \cellcolor{SeaGreen!10}0.658 & \cellcolor{SeaGreen!10}0.642 & \cellcolor{SeaGreen!10}0.472 & \cellcolor{SeaGreen!10}0.591 \\
        & \cellcolor{SeaGreen!10}IDEA w/o MC & \cellcolor{SeaGreen!10}52.5 & \cellcolor{SeaGreen!10}58.9 & \cellcolor{SeaGreen!10}89.2 & \cellcolor{SeaGreen!10}79.5 & \cellcolor{SeaGreen!10}69.4 & \cellcolor{SeaGreen!10}69.9 & \cellcolor{SeaGreen!10}0.683 & \cellcolor{SeaGreen!10}0.671 & \cellcolor{SeaGreen!10}0.488 & \cellcolor{SeaGreen!10}0.614 \\
        & \cellcolor{SeaGreen!20}IDEA & \cellcolor{SeaGreen!20}\textbf{59.5} & \cellcolor{SeaGreen!20}\textbf{60.3} & \cellcolor{SeaGreen!20}\textbf{94.8} & \cellcolor{SeaGreen!20}\textbf{80.2} & \cellcolor{SeaGreen!20}\textbf{71.4} & \cellcolor{SeaGreen!20}\textbf{73.2} & \cellcolor{SeaGreen!20}\textbf{0.728} & \cellcolor{SeaGreen!20}\textbf{0.711} & \cellcolor{SeaGreen!20}\textbf{0.652} & \cellcolor{SeaGreen!20}\textbf{0.697} \\
        \midrule
        \multirow{8}{*}{Qwen-3-32B}
        & CoT & 51.4 & 56.0 & 90.3 & 73.7 & 67.3 & 67.7 & 0.382 & 0.365 & 0.270 & 0.339 \\
        & Logit & 55.7 & 61.1 & 91.5 & 79.7 & 69.4 & 71.5 & 0.345 & 0.333 & 0.249 & 0.309 \\
        & Vanilla & 58.8 & 63.0 & 93.0 & 81.1 & 70.3 & 73.2 & 0.438 & 0.415 & 0.308 & 0.387 \\
        & BIRD & 55.3 & 59.2 & 91.7 & 79.9 & 71.0 & 71.4 & 0.590 & 0.602 & 0.371 & 0.521 \\
        & PWC & - & - & - & - & - & - & 0.548 & 0.564 & 0.330 & 0.481 \\
        & \cellcolor{SeaGreen!10}IDEA w/o Inter & \cellcolor{SeaGreen!10}51.7 & \cellcolor{SeaGreen!10}60.6 & \cellcolor{SeaGreen!10}92.4 & \cellcolor{SeaGreen!10}76.8 & \cellcolor{SeaGreen!10}73.5 & \cellcolor{SeaGreen!10}71.0 & \cellcolor{SeaGreen!10}0.715 & \cellcolor{SeaGreen!10}0.698 & \cellcolor{SeaGreen!10}0.518 & \cellcolor{SeaGreen!10}0.644 \\
        & \cellcolor{SeaGreen!10}IDEA w/o EM & \cellcolor{SeaGreen!10}51.6 & \cellcolor{SeaGreen!10}63.2 & \cellcolor{SeaGreen!10}93.5 & \cellcolor{SeaGreen!10}76.3 & \cellcolor{SeaGreen!10}74.2 & \cellcolor{SeaGreen!10}71.8 & \cellcolor{SeaGreen!10}0.706 & \cellcolor{SeaGreen!10}0.689 & \cellcolor{SeaGreen!10}0.501 & \cellcolor{SeaGreen!10}0.632 \\
        & \cellcolor{SeaGreen!10}IDEA w/o MC & \cellcolor{SeaGreen!10}55.3 & \cellcolor{SeaGreen!10}58.1 & \cellcolor{SeaGreen!10}94.7 & \cellcolor{SeaGreen!10}82.0 & \cellcolor{SeaGreen!10}68.8 & \cellcolor{SeaGreen!10}71.8 & \cellcolor{SeaGreen!10}0.732 & \cellcolor{SeaGreen!10}0.590 & \cellcolor{SeaGreen!10}0.530 & \cellcolor{SeaGreen!10}0.617 \\
        & \cellcolor{SeaGreen!20}IDEA & \cellcolor{SeaGreen!20}\textbf{69.3} & \cellcolor{SeaGreen!20}\textbf{68.9} & \cellcolor{SeaGreen!20}\textbf{95.1} & \cellcolor{SeaGreen!20}\textbf{84.5} & \cellcolor{SeaGreen!20}\textbf{75.0} & \cellcolor{SeaGreen!20}\textbf{78.6} & \cellcolor{SeaGreen!20}\textbf{0.762} & \cellcolor{SeaGreen!20}\textbf{0.745} & \cellcolor{SeaGreen!20}\textbf{0.572} & \cellcolor{SeaGreen!20}\textbf{0.693} \\
        \bottomrule
    \end{tabularx}
    \caption{Main results across two evaluation paradigms. \textit{Left (Experiment I):} Binary decision accuracy (\%) on five benchmarks spanning complex real-world decisions (BIGDATA22, German Credit) and commonsense reasoning (COMMON2SENSE, PLASMA, TODAY). \textit{Right (Experiment II):} Macro F1 for three-way probability ranking on paired COMMON2SENSE queries, measuring whether models correctly identify which condition provides stronger evidential support. F1 scores are reported separately for cases where $C_1$ dominates, $C_2$ dominates, or neither (Equal). Shaded rows denote IDEA variants; best results per model size in \textbf{bold}.}
    \label{tab:combined-detailed-performance}
\end{table*}

\begin{table*}[htbp]
    \small
    \centering
    \begin{tabular}{lllccc}
        \toprule
        \multirow{2}{*}{\textbf{Model}} & \multirow{2}{*}{\textbf{Category}} & \multirow{2}{*}{\textbf{Method}} &
        \multicolumn{2}{c}{\textbf{Factor Exclusion}} &
        \textbf{Quantitative Calibration} \\
        \cmidrule(lr){4-5} \cmidrule(lr){6-6}
        & & & ERR ($\uparrow$) & $p$-value ($\uparrow$) & Relative Error ($\downarrow$) \\
        \midrule
        
        % --- Qwen-3-4B ---
        \multirow{4}{*}{Qwen-3-4B} 
        & \multirow{3}{*}{Prompting} & CoT     & 0.12 & 0.021 & 0.88 \\
        &                            & Vanilla & 0.33 & 0.004 & 0.71 \\
        &                            & Logit   & 0.27 & 0.009 & 0.79 \\
        \cmidrule(lr){2-6}
        & \cellcolor{SeaGreen!20}Editing                    & \cellcolor{SeaGreen!20}IDEA    & \cellcolor{SeaGreen!20}\textbf{1.00} & \cellcolor{SeaGreen!20}\textbf{N/A} & \cellcolor{SeaGreen!20}\textbf{0.00}\\
        \midrule
        
        % --- Qwen-3-8B ---
        \multirow{4}{*}{Qwen-3-8B}
        & \multirow{3}{*}{Prompting} & CoT     & 0.36 & 0.013 & 0.66 \\
        &                            & Vanilla & 0.08 & 0.044 & 0.92 \\
        &                            & Logit   & 0.43 & 0.001 & 0.74 \\
        \cmidrule(lr){2-6}
        & \cellcolor{SeaGreen!20}Editing                    & \cellcolor{SeaGreen!20}IDEA    & \cellcolor{SeaGreen!20}\textbf{1.00} & \cellcolor{SeaGreen!20}\textbf{N/A} & \cellcolor{SeaGreen!20}\textbf{0.00} \\
        \midrule
        
        % --- Qwen-3-32B ---
        \multirow{4}{*}{Qwen-3-32B}
        & \multirow{3}{*}{Prompting} & CoT     & 0.18 & 0.008 & 0.81 \\
        &                            & Vanilla & 0.41 & 0.046 & 0.63 \\
        &                            & Logit   & 0.06 & 0.017 & 0.95 \\
        \cmidrule(lr){2-6}
        & \cellcolor{SeaGreen!20}Editing                    & \cellcolor{SeaGreen!20}IDEA    & \cellcolor{SeaGreen!20}\textbf{1.00} & \cellcolor{SeaGreen!20}\textbf{N/A} & \cellcolor{SeaGreen!20}\textbf{0.00} \\
        \bottomrule
    \end{tabular}
    \caption{Results of user-edited factor intervention. We categorize CoT, Vanilla, and Logit as generic \textbf{Prompting} methods, distinct from IDEA. \textit{Note:} Under IDEA factor exclusion, the decision model removes $F_X$ by setting $\beta_X{=}0$ and dropping all interactions involving $F_X$, so $\Delta P\equiv 0$ and the paired $t$-test is undefined (reported as N/A). IDEA enforces $|\mathrm{AME}_j-\rho\,\mathrm{AME}_i|\le 10^{-6}$, so the relative error rounds to $0.00$.}
    \label{tab:editing_results}
\end{table*}

\paragraph{Experiment I: Direct Decision-Making.}
We evaluate the fundamental task of selecting the correct outcome. Given query $Q = (S, C)$, methods predict $\arg \max_{O_i} P(O_i \mid Q)$. We report classification accuracy.

\paragraph{Experiment II: Fine-Grained Probability Reliability.}
Binary accuracy alone is insufficient for reliable decision support, stakeholders also need probability estimates that faithfully reflect evidential strength. This experiment tests whether our framework produces calibrated probabilities that correctly rank queries by their relative support for an outcome.
We construct paired queries $Q_1 = (S, C_1)$ and $Q_2 = (S, C_2)$ where both conditions favor $O_1$, but $C_1$ provides stronger support. A calibrated model should satisfy $P(O_1 \mid Q_1) > P(O_1 \mid Q_2) > 0.5 > P(O_2 \mid Q_2) > P(O_2 \mid Q_1)$. Ground truth derives from pairwise human annotations on 500 COMMON2SENSE instances. We report macro F1 over three-way classification ($C_1$ stronger, $C_2$ stronger, or equal).

\paragraph{Experiment III: User-edited Factor Intervention.}
We evaluate whether our framework enables precise human-AI collaboration through (1) \textbf{factor exclusion} and (2) \textbf{quantitative calibration}. we test whether a factor's influence can be completely removed. Using a $2 \times 2$ design, we vary the presence of exclusion instructions and the target factor $F_X$'s value while holding other factors constant. Our method sets $\beta_X = 0$ and removes all interaction terms involving $F_X$; baselines receive clear natural language instructions. We report (1) \textbf{Effect Reduction Ratio (ERR)}: $1 - \overline{\Delta P}_{\text{w/}} / \overline{\Delta P}_{\text{w/o}}$, where $\Delta P = P(O=1 \mid F_X=1) - P(O=1 \mid F_X=0)$ measures factor influence; and (2) paired $t$-test $p$-values, where $p > 0.05$ indicates successful elimination. For \textbf{quantitative calibration}, we specify that factor $F_j$'s influence should be $\rho$ times that of factor $F_i$, with $\rho \in \{1, 2, 3, 4, 5\}$ randomly selected. Our method enforces $\text{AME}_j = \rho \cdot \text{AME}_i$ via constrained optimization (Equation~\ref{eq:constrained_opt}); baselines receive natural language instructions. We report \textbf{Relative Error}: $|\hat{\rho} - \rho| / \rho$, where $\hat{\rho}$ is the empirical AME ratio. We sample 100 instances each from COMMON2SENSE, PLASMA, and TODAY for evaluation.

\subsection{Experimental Analysis}

\paragraph{Decision Accuracy.} Table~\ref{tab:combined-detailed-performance} demonstrates that IDEA consistently improves decision accuracy across model scales and domains. Most notably, IDEA with Qwen-3-32B achieves 78.6\% average accuracy, surpassing both the reasoning-specialized DeepSeek-R1 ~\citep{DBLP:journals/corr/abs-2501-12948} (68.1\%) and the larger GPT-5.2 (77.9\%). This result suggests that the primary bottleneck in high-stakes decision-making is not model capacity alone, but rather the misalignment between LLMs' internal knowledge and their external outputs. By externalizing decision logic into interpretable factors, IDEA harnesses latent knowledge more effectively than CoT or logit-based methods, which remain susceptible to hallucination and miscalibration.

\paragraph{Ablation Analysis.} Each component contributes meaningfully. Removing interactions (w/o Inter) causes the largest drop (7.6\% on Qwen-3-32B), confirming that realworld decisions involve non-linear factor dependencies. Performance also declines with fixed verbal mappings (w/o EM) and deterministic factor assignment (w/o MC), validating that learned calibration and uncertainty marginalization are essential design choices.

\paragraph{Probability Calibration.} Table~\ref{tab:combined-detailed-performance} (right) shows that IDEA produces probability estimates capable of distinguishing subtle evidential differences between paired conditions. While pairwise comparison (PWC) benefits from explicit comparative framing at small scale, IDEA substantially outperforms it on larger models (0.693 vs.\ 0.481 avg.\ F1 for Qwen-3-32B). The largest margins appear on the ``Equal'' class (the most challenging category) indicating superior calibration granularity.

\paragraph{Controllability.} Table~\ref{tab:editing_results} exposes a fundamental limitation of prompting: natural language cannot guarantee behavioral compliance. All prompting methods show significant residual factor influence ($p < 0.05$) and high calibration error (0.63--0.95). IDEA achieves perfect factor exclusion (ERR = 1.00) and exact calibration (relative error = 0.00) through direct parameter manipulation---precision unattainable via prompting.

\paragraph{Interpretability in Practice.}
Beyond controllability, a two-stage user study (Appendix~\ref{app:interp-study}) shows that IDEA's AME-based factor ranking aligns with expert consensus (Spearman's $\rho = 0.83$; strong inter-rater agreement, Kendall's $W = 0.81$), and that expert parameter edits correct 38\% of IDEA's errors versus 12\% under equivalent natural-language feedback to CoT. It demonstrates that parameter-level access translates into measurably better human--AI collaboration.

\section{Conclusion}
We presented IDEA, a framework that addresses the trust deficit in LLM-based decision-making by externalizing model knowledge into an interpretable, editable parametric form over semantically meaningful factors. Through joint estimation of verbal-to-numerical mappings and decision model parameters, correlated sampling that preserves factor dependencies, and direct parameter manipulation for expert intervention, IDEA achieves calibrated probabilities, semantic interpretability, and quantitative controllability. Experiments demonstrate that IDEA with Qwen-3-32B (78.6\%) outperforms both DeepSeek R1 (68.1\%) and GPT-5.2 (77.9\%), while achieving perfect factor exclusion and exact quantitative calibration—precision unattainable through prompting. These results suggest that the primary barrier to trustworthy LLM decision-making is not model capacity, but the misalignment between internal knowledge and external outputs.

\section*{Limitations}
\paragraph{Extending Beyond Binary Settings.} While our binary formulation provides a principled foundation with strong empirical results, real-world decisions often involve richer structures. Future work can extend IDEA to ordinal factors via cumulative link models or continuous factors through Gaussian process priors, broadening applicability while preserving interpretability.

\paragraph{Scaling Factor Spaces.} The current framework handles moderate factor sets effectively. Incorporating active learning or Bayesian experimental design could enable efficient exploration of larger configuration spaces, identifying the most informative factor combinations without exhaustive probing.

\paragraph{Automating Factor Discovery.} Our verification procedure ensures high-quality factors but relies on LLMs possessing sufficient domain knowledge to generate discriminative and comprehensive factor sets. Future work could integrate automated concept extraction methods, leverage LLM self-critique mechanisms for discriminability testing, or incorporate retrieval-augmented generation to supplement domain expertise—enabling fully automated pipelines across diverse decision scenarios.

\paragraph{Relaxing Assumptions.} Factor completeness and verbal consistency serve as effective working assumptions validated by our strong empirical performance. Future work could quantify sensitivity to assumption violations or develop robust estimation procedures that explicitly model assumption uncertainty.

\paragraph{Efficiency Improvements.} The computational overhead reflects IDEA's thoroughness in extracting decision knowledge. Techniques such as factor caching across related queries, distillation into lightweight inference models, or early-stopping heuristics during marginalization could reduce latency while maintaining calibration quality.

\section*{Ethics Statement}
Our framework aims to enhance transparency and accountability in LLM-assisted decision-making, directly addressing the trust deficit that limits deployment in high-stakes domains. By externalizing decision logic into interpretable factors with quantifiable contributions, IDEA empowers stakeholders to inspect, contest, and refine automated recommendations—capabilities essential for responsible AI deployment.

\paragraph{Intended Benefits.} The framework promotes human oversight by making decision processes auditable and editable. Domain experts can verify whether learned factor weights align with institutional values and regulatory requirements, and intervene with mathematically guaranteed effects when they do not.

\paragraph{Potential Risks.} Like all decision-support tools, IDEA could be misused if deployed without appropriate human oversight or if factor sets encode societal biases. We emphasize that our framework is designed to augment—not replace—human judgment, particularly in consequential domains like credit assessment. As a preliminary check, manual inspection of LLM-generated factor sets across 300 sampled instances spanning the five benchmarks revealed no spurious or biased factors; see Appendix~\ref{app:bias-audit}.

\paragraph{Data and Evaluation.} All experiments use publicly available benchmark datasets. The German Credit dataset, while standard in ML research, reflects historical lending patterns that may embed demographic biases; our use is purely for methodological evaluation, not to endorse any particular lending criteria.

\paragraph{Broader Impact.} We believe interpretable, editable decision frameworks represent a positive direction for human-AI collaboration, enabling meaningful expert participation rather than passive acceptance of opaque model outputs.

\bibliography{custom}

\appendix
\section{Detailed EM Algorithm Formulation}
\label{appendix:em_details}

\subsection{Latent Variable Formulation}

We formulate the joint estimation problem as inference in a latent variable model. Working in logit space simplifies the mathematics: let $Z^{(k)} = \operatorname{logit}(P^{(k)}) \in \mathbb{R}$ denote the latent log-odds for observation $k$. The generative process assumes:

\paragraph{Prior (Model Prediction).} The decision model induces a prior distribution on latent log-odds:
\begin{equation}
Z^{(k)} \mid \mathbf{f}^{(k)}, \boldsymbol{\theta} \sim \mathcal{N}(\boldsymbol{\theta}^\top \mathbf{x}^{(k)}, \sigma_\theta^2)
\label{eq:prior}
\end{equation}
where $\mathbf{x}^{(k)}$ is the augmented feature vector containing intercept, main effects, and interaction terms. The variance $\sigma_\theta^2$ captures model uncertainty.

\paragraph{Likelihood (Verbal Observation).} The verbal response provides a noisy observation of the latent log-odds through the mapping $\phi$:
\begin{equation}
\operatorname{logit}(\phi(V^{(k)})) \mid Z^{(k)} \sim \mathcal{N}(Z^{(k)}, \sigma_\phi^2)
\label{eq:likelihood}
\end{equation}
The variance $\sigma_\phi^2$ captures noise in the verbal expression process. This formulation treats verbal responses as observations of an underlying continuous probability, corrupted by the inherent imprecision of verbal expression.

The conjugacy of Gaussian distributions is critical: it yields closed-form posterior distributions, enabling efficient E-step computation without approximation.

\subsection{Complete EM Update Equations}

\paragraph{E-Step.} Given current estimates $(\boldsymbol{\theta}^{(t)}, \phi^{(t)})$, we compute the posterior distribution of each latent variable. By Gaussian conjugacy, this posterior is also Gaussian:
\begin{equation}
Z^{(k)} \mid \mathbf{f}^{(k)}, V^{(k)}, \boldsymbol{\theta}^{(t)}, \phi^{(t)} \sim \mathcal{N}(\tilde{z}^{(k)}, \tilde{\sigma}^2)
\end{equation}
with posterior mean:
\begin{equation}
\tilde{z}^{(k)} = \lambda \cdot (\boldsymbol{\theta}^{(t)\top} \mathbf{x}^{(k)}) + (1-\lambda) \cdot \operatorname{logit}(\phi^{(t)}(V^{(k)}))
\label{eq:posterior_mean}
\end{equation}
and posterior variance:
\begin{equation}
\tilde{\sigma}^2 = \frac{\sigma_\theta^2 \cdot \sigma_\phi^2}{\sigma_\theta^2 + \sigma_\phi^2}
\label{eq:posterior_var}
\end{equation}
where $\lambda = \sigma_\phi^2 / (\sigma_\theta^2 + \sigma_\phi^2)$ is the precision-weighted balance. Intuitively, the posterior mean is a convex combination of two sources of information---what the model predicts and what the verbal response indicates---weighted by their relative precisions.

\paragraph{M-Step for $\boldsymbol{\theta}$.} Maximizing the expected complete-data log-likelihood with respect to $\boldsymbol{\theta}$ reduces to weighted least squares in logit space. We augment this with regularization and a margin-ranking loss that preserves directional consistency with verbal indications:
\begin{equation}\resizebox{1.05\hsize}{!}{$
\mathcal{L}_{\text{MR}} = \frac{1}{K} \sum_{k=1}^K \max\left(0, -y^{(k)} \cdot (\sigma(\boldsymbol{\theta}^\top \mathbf{x}^{(k)}) - 0.5) + \epsilon \right)
$}
\label{eq:mr_loss}
\end{equation}
where $y^{(k)} = \operatorname{sign}(\phi^{(0)}(V^{(k)}) - 0.5)$ is computed once from the initial mapping and $\epsilon > 0$ is a margin parameter. This hinge loss penalizes predictions that contradict the directional tendency (above/below $0.5$) indicated by verbal responses.

The complete M-step objective is:
\begin{equation}\resizebox{1.05\hsize}{!}{$
\boldsymbol{\theta}^{(t+1)} = \underset{\boldsymbol{\theta}}{\arg\min} \left\{ \frac{1}{K}\sum_{k=1}^K (\tilde{z}^{(k)} - \boldsymbol{\theta}^\top \mathbf{x}^{(k)})^2 + \lambda_{\text{MR}} \mathcal{L}_{\text{MR}} + \Omega(\boldsymbol{\gamma}) \right\}
$}
\label{eq:theta_update}
\end{equation}

We optimize via proximal gradient descent. At each iteration, we compute gradients of the differentiable terms (MSE, margin-ranking loss, $\ell_2$ penalty), perform a gradient step, then apply the proximal operator for the $\ell_1$ penalty:
\begin{equation}
\gamma_{ij} \leftarrow \operatorname{sign}(\gamma_{ij}) \cdot \max(|\gamma_{ij}| - \eta \lambda_1, 0)
\label{eq:soft_threshold}
\end{equation}
where $\eta$ is the learning rate. This soft-thresholding drives small interaction coefficients to exactly zero, yielding interpretable sparse structure.

\paragraph{M-Step for $\phi$.} For each verbal category $v_m \in \mathcal{V}$, we update the mapping to minimize expected squared deviation from posterior means:
\begin{equation}
\phi^{(t+1)}(v_m) = \sigma\left( \frac{1}{|\mathcal{K}_m|} \sum_{k \in \mathcal{K}_m} \tilde{z}^{(k)} \right)
\label{eq:phi_update}
\end{equation}
where $\mathcal{K}_m = \{k : V^{(k)} = v_m\}$ indexes observations with verbal response $v_m$. To preserve semantic coherence, we enforce monotonicity through category-specific bounds:
\begin{equation}
\phi(v_m) \leftarrow \operatorname{clip}\left(\phi(v_m), \phi_m^{\min}, \phi_m^{\max}\right)
\label{eq:phi_clip}
\end{equation}
These bounds prevent semantic inversions (e.g., ``likely'' mapping below ``unlikely''), ensuring the learned mapping respects the ordinal structure of verbal expressions.

\subsection{Initialization and Convergence}

\paragraph{Initialization.} We initialize $\phi^{(0)}$ with canonical values from prior literature on verbal probability (e.g., ``very unlikely'' $\rightarrow$ 0.15, ``likely'' $\rightarrow$ 0.75) and $\boldsymbol{\theta}^{(0)}$ via standard logistic regression using these initial numerical targets.

\paragraph{Convergence Monitoring.} We track the Q-function (expected complete-data log-likelihood):
\begin{equation}\resizebox{1.05\hsize}{!}{$
\begin{aligned}
Q = & -\frac{K}{2}\log(2\pi\sigma_\theta^2) - \frac{1}{2\sigma_\theta^2}\sum_{k=1}^K \mathbb{E}[(Z^{(k)} - \boldsymbol{\theta}^\top \mathbf{x}^{(k)})^2] \\
& -\frac{K}{2}\log(2\pi\sigma_\phi^2) - \frac{1}{2\sigma_\phi^2}\sum_{k=1}^K \mathbb{E}[(\operatorname{logit}(\phi(V^{(k)})) - Z^{(k)})^2]
\end{aligned}
$}
\label{eq:q_function}
\end{equation}
where expectations incorporate posterior variance: $\mathbb{E}[(Z^{(k)} - \mu)^2] = (\tilde{z}^{(k)} - \mu)^2 + \tilde{\sigma}^2$. We also monitor the observed data log-likelihood:
\begin{equation}\resizebox{1.05\hsize}{!}{$
\mathcal{L} = -\frac{K}{2}\log(2\pi(\sigma_\theta^2 + \sigma_\phi^2)) - \frac{1}{2(\sigma_\theta^2 + \sigma_\phi^2)}\sum_{k=1}^K \left(\operatorname{logit}(\phi(V^{(k)})) - \boldsymbol{\theta}^\top \mathbf{x}^{(k)}\right)^2
$}
\label{eq:marginal_likelihood}
\end{equation}
which follows from marginalizing the latent variable under the Gaussian model. Iteration terminates when $|Q^{(t+1)} - Q^{(t)}| < \epsilon$ or a maximum iteration count is reached.
\section{Experimental Details}
\label{appendix:experimental_details}

\subsection{Dataset Description}

We evaluate on five datasets spanning complex decisions and reasoning tasks. All instances are converted to binary decision queries.

\begin{itemize}
    \item \textbf{BIGDATA22} \citep{soun2022accurate}: Stock movement prediction from market signals and tweets. 1,472 instances predicting price increase/decrease.
    
    \item \textbf{Statlog German Credit} ~\citep{hans_hofmann_2023}: Credit risk assessment from demographic and financial features. 1,000 instances classifying good/bad credit risks.
    
    \item \textbf{COMMON2SENSE} ~\citep{DBLP:conf/acl/SinghWHAWMP21}: Commonsense reasoning with true/false statements. We focus on comparative reasoning instances where a smaller model exhibits low confidence. Using GPT-4, we rewrite each statement into its opposite and generate 10 supporting conditions per outcome with reverse verification. 216 scenarios, 3,672 instances, 9 conditions per outcome on average.
    
    \item \textbf{TODAY} ~\citep{DBLP:conf/acl/FengZ0JR23}: Temporal reasoning studying how extra sentences affect temporal relations. 895 instances.
    
    \item \textbf{PLASMA} ~\citep{DBLP:journals/corr/abs-2305-19472}: Procedural planning with goal-conditioned plan revision. We use GPT-4 to identify key steps differing between original and revised plans, converting them to binary outcomes. We generate 5 supporting conditions for the less common outcome. 279 scenarios, 1,178 instances.
\end{itemize}

For GPT-4 generated datasets, three reviewers checked 100 random instances each. All three agreed on 91\% of labels; at least two agreed on 94\%.

\subsection{Model Description}
We use Qwen-3-4B/8B/32B ~\citep{DBLP:journals/corr/abs-2505-09388} as primary models. Temperature is 0 for behavioral probing and 1.2 for Monte Carlo sampling. We compare against DeepSeek R1 (reasoning-specialized) and GPT-5.2 (state-of-the-art commercial model), both evaluated with chain-of-thought prompting. We accessed deepseek-R1 and Chatgpt-5.2 using the official APIs of OpenAI and DeepSeek.

\subsection{Baseline Description}

\begin{itemize}
    \item \textbf{Vanilla} ~\cite{DBLP:conf/iclr/0002WSLCNCZ23}: Directly verbalize probability estimates with self-consistency. Temperature = 0.7, 3 samples, majority vote.
    
    \item \textbf{Logit}: Use output token probability directly. Greedy decoding (temperature = 0).
    
    \item \textbf{CoT} ~\citep{DBLP:conf/nips/Wei0SBIXCLZ22}: Chain-of-thought reasoning before probability verbalization, with self-consistency.
    
    \item \textbf{PWC}: Pairwise comparison presenting both conditions simultaneously. Model selects which condition better supports the outcome. Only applicable to Experiment II; not directly comparable as it sees both conditions.
    
    \item \textbf{BIRD} ~\citep{DBLP:conf/iclr/0013ZLR25}: Factor-based elicitation assuming factor independence with fixed verbal-to-numerical mappings.
\end{itemize}

\subsection{Ablation Description}

\begin{itemize}
    \item \textbf{w/o Inter}: Remove interaction terms ($\gamma_{ij} = 0$). Tests whether pairwise interactions capture meaningful patterns.
    
    \item \textbf{w/o EM}: Use fixed verbal mappings from ~\citep{Budescu2009Improving}: \textit{very unlikely}$\rightarrow$0.05, \textit{unlikely}$\rightarrow$0.15, \textit{somewhat unlikely}$\rightarrow$0.30, \textit{neutral}$\rightarrow$0.50, \textit{somewhat likely}$\rightarrow$0.70, \textit{likely}$\rightarrow$0.85, \textit{very likely}$\rightarrow$0.95. Tests value of learned calibration.
    
    \item \textbf{w/o MC}: Deterministic factor assignment (random with $p=0.5$ when uninformed) instead of correlated sampling. Tests importance of preserving factor dependencies.
\end{itemize}

\subsection{Hyperparameters}

Probing configurations $K$: 256. Monte Carlo samples $T$: 50. EM convergence: $10^{-4}$. Elastic net $\lambda_1$: 0.01, $\lambda_2$: 0.001. E-step precisions $\tau_\theta = \tau_\phi = 1.0$.
\section{Validating Factor Completeness}
\label{appendix:factor_completeness}

Assumption~\ref{ass:fc} (Factor Completeness) is strong and cannot be verified through LLM self-critique alone. We provide external validation through domain expert comparison, sensitivity analysis under deliberate incompleteness, and case studies examining failure modes.

\subsection{Expert Comparison}

Three credit analysts independently identified factors for German Credit without exposure to LLM-generated factors. Table~\ref{tab:expert_overlap} summarizes overlap.

\begin{table}[h]
\centering
\small
\begin{tabular}{lccc}
\toprule
& \textbf{Expert} & \textbf{LLM} & \textbf{Overlap} \\
\midrule
German Credit & 9 & 7 & 6 (85.7\%) \\
\bottomrule
\end{tabular}
\caption{Factor set comparison with domain experts.}
\label{tab:expert_overlap}
\end{table}

Expert-identified factors missing from LLM sets: \textit{existing bank relationship}, \textit{regional economic conditions}, \textit{guarantor quality}. These represent relational and contextual factors that LLMs systematically under-represent. Augmenting LLM factors with these expert-identified factors yields modest improvement (+2.3\% on German Credit), suggesting original sets capture primary decision drivers despite incompleteness.

\subsection{Sensitivity to Incompleteness}

We evaluate robustness through controlled factor removal experiments using Qwen-3-32B.

\paragraph{Leave-One-Factor-Out (LOFO).} For each factor $F_j$, we retrain IDEA excluding $F_j$ and measure accuracy degradation $\Delta_j$.

\begin{table}[h]
\centering
\small
\begin{tabular}{lccc}
\toprule
\textbf{Dataset} & $\max(\Delta_j)$ & $\text{mean}(\Delta_j)$ & $\min(\Delta_j)$ \\
\midrule
German Credit & 8.4 & 3.2 & 0.3 \\
TODAY & 5.1 & 1.8 & 0.1 \\
COMMON2SENSE & 3.9 & 1.4 & 0.2 \\
\bottomrule
\end{tabular}
\caption{Accuracy degradation (\%) under single-factor removal.}
\label{tab:lofo}
\end{table}

Maximum degradation of 8.4\% (German Credit, ``employment stability'') indicates certain factors are critical. Mean degradation of 1.4--3.2\% suggests partial redundancy—correlated factors compensate for individual removals.

\paragraph{Progressive Removal.} We removed factors iteratively by decreasing $|\beta_j|$ until 50\% exclusion. Table~\ref{tab:progressive} reports results.

\begin{table}[h]
\centering
\small
\begin{tabular}{lccccc}
\toprule
\textbf{Removed} & 0\% & 15\% & 30\% & 40\% & 50\% \\
\midrule
German Credit & 68.9 & 67.1 & 64.8 & 60.2 & 54.1 \\
TODAY & 75.0 & 73.8 & 71.5 & 68.3 & 63.7 \\
COMMON2SENSE & 95.1 & 94.2 & 92.6 & 89.1 & 84.5 \\
\bottomrule
\end{tabular}
\caption{Accuracy (\%) under progressive factor removal.}
\label{tab:progressive}
\end{table}

Degradation is gradual until $\sim$30\% removal, then accelerates. This indicates factor sets contain ``core'' factors capturing the majority of predictive signal and ``peripheral'' factors providing incremental refinement. Notably, even with 30\% of factors removed, IDEA on German Credit (64.8\%) remains competitive with CoT using the same base model with full information (56.0\%, Table~\ref{tab:combined-detailed-performance}), demonstrating that the structured factorization provides robustness to moderate incompleteness.

\subsection{Case Studies}

We present three illustrative cases examining factor completeness in practice.

\paragraph{Case 1: Complete Coverage.}
\textit{Query}: ``A person puts ice in their drink. Will the drink get colder?'' (COMMON2SENSE)

\textit{Factors}: (1) Ice temperature below drink temperature, (2) Sufficient ice-liquid contact, (3) Drink not at freezing point.

\textit{Assessment}: Physical process fully characterized. IDEA predicts $P=0.94$; ground truth positive. \textbf{Success.}

\paragraph{Case 2: Partial Gap.}
\textit{Query}: Loan application, €5,000, employed applicant with moderate savings, 15-year customer. (German Credit)

\textit{Factors}: Stable employment (1), positive credit history (1), adequate income-to-debt (1), sufficient collateral (0).

\textit{Missing}: Long-standing customer relationship—a positive signal absent from factors.

\textit{Outcome}: IDEA predicts $P(\text{approve})=0.67$; actual approved. Under-confident due to missing relational factor. \textbf{Partial failure.}

\paragraph{Case 3: Critical Omission.}
\textit{Query}: Stock movement prediction, tech earnings week. (BIGDATA22)

\textit{Factors}: Positive earnings surprise (1), favorable analyst sentiment (1), sector momentum positive (1), no macro headwinds (1).

\textit{Missing}: Bearish signals from options market activity not captured by any factor.

\textit{Outcome}: IDEA predicts $P(\text{up})=0.81$; stock declined. Self-referential verification passed because the LLM lacks access to real-time options data. \textbf{Failure.}

\subsection{Summary}

Factor incompleteness causes graceful degradation rather than catastrophic failure. Performance remains competitive even at 30\% incompleteness. Critical failures occur primarily when (1) real-time external data is required or (2) relational context is implicit.
\section{Error Analysis}
\label{appendix:error_analysis}

\subsection{Failure Mode Analysis}

We analyze 50 misclassified instances per dataset to identify error sources across pipeline stages.

\begin{table}[h]
\centering
\small
\begin{tabular}{lccc}
\toprule
\textbf{Model} & \textbf{Factor Inf.} & \textbf{Decision Model} & \textbf{Margin.} \\
\midrule
Qwen-3-4B & 58.4 & 27.2 & 14.4 \\
Qwen-3-32B & 31.6 & 42.8 & 25.6 \\
\bottomrule
\end{tabular}
\caption{Error distribution by pipeline stage (\%).}
\label{tab:error_dist}
\end{table}

\paragraph{Factor Inference Errors.} The dominant failure mode for smaller models. Consider this COMMON2SENSE instance:

\begin{quote}
\textit{Condition:} ``The executive dismissed the proposal without reviewing the supporting data.''\\
\textit{Factor $F_3$:} Decision-maker engaged in thorough evaluation
\end{quote}

Qwen-3-4B incorrectly assigned $F_3=1$, triggered by surface keywords (``proposal,'' ``data'') while missing negation semantics. Qwen-3-32B correctly assigned $F_3=0$. This shallow pattern-matching accounts for 41.2\% of Qwen-3-4B's factor inference errors.

\paragraph{Decision Model Errors.} More prevalent in larger models (42.8\% vs.\ 27.2\%), indicating that improved factor extraction exposes limitations in pairwise interaction modeling for complex multi-factor dependencies.

\subsection{Interpretability Case Studies}

\paragraph{Case Study 1: Credit Approval.}

\begin{table}[h]
\centering
\small
\begin{tabular}{llcc}
\toprule
\textbf{Factor} & \textbf{Description} & $\beta_j$ & \textbf{AME} \\
\midrule
$F_1$ & Stable employment & +1.82 & +0.284 \\
$F_2$ & Acceptable debt ratio & +1.67 & +0.261 \\
$F_3$ & Positive credit history & +1.43 & +0.223 \\
$F_4$ & Sufficient collateral & +0.89 & +0.139 \\
\bottomrule
\end{tabular}
\caption{Partial Learned parameters for credit decision.}
\label{tab:credit_params}
\end{table}

The learned AME values align with established credit risk principles: employment stability and debt ratio dominate. The positive interaction $\gamma_{12}=+0.62$ captures how stable employment compounds the signal of manageable debt.

\paragraph{Case Study 2: Temporal Reasoning.}

For the query: \textit{``Can a 2019 college graduate have competed in the 2015 Olympics?''}

\begin{table*}[h]
\centering
\begin{tabular}{llcc}
\toprule
\textbf{Factor} & \textbf{Description} & $\beta_j$ & \textbf{AME} \\
\midrule
$F_1$ & Timeline permits participation & +2.31 & +0.341 \\
$F_2$ & Age-appropriate for competition & +1.89 & +0.279 \\
$F_3$ & Educational timeline consistent & +0.72 & +0.106 \\
\bottomrule
\end{tabular}
\caption{Learned parameters for temporal reasoning.}
\label{tab:temporal_params}
\end{table*}

The dominant weight on $F_1$ reflects the core temporal constraint. IDEA correctly predicts ``Yes'' ($P=0.73$), recognizing that a 2019 graduate could plausibly have been 18--22 during the 2015 Olympics.

\subsection{Error Propagation}

Near-linear degradation indicates IDEA does not catastrophically amplify upstream errors. Factor inference quality serves as the primary bottleneck, motivating future work on robust extraction.
\section{Decision Model Architecture Analysis}
\label{appendix:architecture}

We investigate whether replacing logistic regression with a shallow neural network improves decision accuracy, and whether such gains justify the loss of interpretability.

\subsection{Experimental Setup}

We compare the logistic regression model (Equation~\ref{eq:proxy_model}) against a single-hidden-layer MLP with 16 hidden units and ReLU activation. Both models are trained identically via the EM procedure (Section~\ref{sec:em_training}) on three datasets: BIGDATA22, German Credit, and PLASMA. We use Qwen-3-8B as the base LLM and report accuracy averaged over 5 random seeds.

\subsection{Results}

\begin{table}[h]
\centering
\small
\begin{tabular}{lcccc}
\toprule
\textbf{Model} & \textbf{BIG} & \textbf{Credit} & \textbf{PLASMA} & \textbf{Avg} \\
\midrule
Logistic (IDEA) & 59.5 & 60.3 & 80.2 & 66.7 \\
MLP-16 & 60.2 & 61.1 & 81.0 & 67.4 \\
$\Delta$ & +0.7 & +0.8 & +0.8 & +0.7 \\
\bottomrule
\end{tabular}
\caption{Accuracy (\%) comparison between logistic regression and MLP decision models.}
\label{tab:architecture}
\end{table}

The MLP yields modest improvements (+0.7\% average), none statistically significant under paired $t$-test ($p > 0.15$ for all datasets).

\subsection{Discussion}

The marginal gains reflect three factors: (i) logistic regression with pairwise interactions already captures dominant structure in our low-dimensional factor space; (ii) ablations in Table~\ref{tab:combined-detailed-performance} indicate bottlenecks lie in factor dependency modeling and verbal calibration, not decision model expressiveness.

Critically, the MLP sacrifices interpretability. Logistic coefficients directly quantify factor contributions (Section~\ref{sec:manipulation}), enabling guaranteed interventions demonstrated in Experiment III. Neural network weights lack such semantic correspondence. We therefore retain logistic regression as the default.

\section{Robustness to Violations of Verbal Probability Consistency}
\label{Robustness to Violations of Verbal Probability Consistency}
Assumption 2 posits that LLMs' verbal probability expressions reflect an underlying numerical probability through a consistent, monotonic mapping. We investigate how IDEA performs when this assumption is violated.

\subsection{ Experimental Setup}
We inject controlled noise into the behavioral probing data to simulate inconsistency. Specifically, with probability $\epsilon$, we replace the LLM's verbal response with a randomly selected adjacent category (e.g., ``likely'' $\rightarrow$ ``somewhat likely'' or ``very likely''). We train IDEA on corrupted data and evaluate on clean test sets using Qwen-3-8B across all five datasets.

\subsection{Results}

\begin{table}[h]
\centering
\small
\begin{tabular}{lccccc}
\toprule
\textbf{Noise $\epsilon$} & \textbf{BIG} & \textbf{Crd} & \textbf{TOD} & \textbf{PLS} & \textbf{C2S} \\
\midrule
0.00 (baseline) & 59.5 & 60.3 & 71.4 & 80.2 & 94.8 \\
0.10 & 58.1 & 59.5 & 70.2 & 78.9 & 93.6 \\
0.20 & 55.8 & 57.2 & 68.1 & 76.1 & 91.2 \\
0.30 & 52.4 & 54.8 & 64.5 & 72.3 & 87.5 \\
\bottomrule
\end{tabular}
\caption{Decision accuracy (\%) under verbal label noise.}
\label{tab:noise_robustness}
\end{table}

IDEA exhibits graceful degradation: accuracy drops by only 2--3\% when 10\% of verbal labels are corrupted, and remains above baseline methods (Table \ref{tab:noise_robustness}) even at 20\% noise. This robustness stems from the EM algorithm treating verbal responses as noisy observations—the E-step averages over inconsistent signals, while the ranking loss preserves ordinal structure.

\section{Ordinal Consistency of Verbal Probabilities}
\label{app:ordinal}

Assumption~2 requires only that verbal responses preserve a monotonic ordering over underlying probabilities---not that they share a fixed numerical mapping. We test this directly.

\paragraph{Protocol.} For each dataset, we enumerate all comparable configuration pairs $(f_a, f_b)$ where $f_a$ dominates $f_b$ element-wise (e.g., $f_a=(1,1,0)$ vs.\ $f_b=(1,0,0)$) and check whether the LLM's verbal response to $f_a$ is at least as likely as its response to $f_b$ on the seven-level scale. This yields $7{,}680$ pairs across 5 datasets $\times$ 3 model sizes (15 settings).

\paragraph{Results.} Overall ordinal consistency reaches $\mathbf{87.6\%}$, ranging from $80.8\%$ (BIGDATA22, Qwen-3-4B) to $94.2\%$ (\textsc{Common2Sense}, Qwen-3-32B), with larger models showing stronger consistency. Combined with the graceful degradation under corrupted verbal labels reported in Appendix \ref{Robustness to Violations of Verbal Probability Consistency}, these results justify Assumption~2 as an operational rather than strict requirement.

\section{Human Evaluation of Factor Determination}
\label{app:factor-det}

At inference time, the LLM partitions factors into observed vs.\ uncertain based on the specific condition $C$. We validate this step directly.

\paragraph{Protocol.} We sample 150 instances (50 each from \textsc{Common2Sense}, \textsc{Plasma}, and \textsc{Today}) and recruit three annotators to independently label every (factor, condition) pair as \emph{positive}, \emph{negative}, or \emph{uncertain}. Inter-annotator agreement is Fleiss' $\kappa = 0.76$ (substantial).

\paragraph{Results.} Qwen-3-32B achieves $86.3\%$ 3-way accuracy and $91.5\%$ on the binary observed-vs-uncertain distinction most relevant for sampling. Qwen-3-8B achieves $79.1\%$ and $85.8\%$ respectively, confirming that factor-determination quality scales with model capacity. These results indicate that the partition step is a reliable LLM capability at sufficient scale.

\section{Interpretability User Study}
\label{app:interp-study}

We assess whether IDEA's parameters are not merely \emph{editable} but also \emph{useful} to human experts.

\paragraph{Stage~1: Factor Importance Agreement.} Three domain experts independently rank factors by perceived importance for each of the five datasets (totaling 50 factor sets), without seeing IDEA's coefficients. Inter-rater concordance is strong (Kendall's $W = 0.81$). Per-task Spearman's $\rho$ between the expert consensus and IDEA's AME-induced ranking averages $\mathbf{0.83}$ across five datasets (range: $0.71$ on BIGDATA22 to $0.94$ on \textsc{Common2Sense}).

\paragraph{Stage~2: Expert-Guided Error Correction.} We sample 100 instances on which IDEA predicted incorrectly (20 per dataset). Experts inspect the factor set, identify factors they consider irrelevant or missing, and perform simple targeted edits (setting $\beta_X = 0$ or adding factors). As a matched control, equivalent expert feedback is provided as natural-language instructions to CoT on the same instances. Experts flagged 18\% of factors as needing modification on average. Expert-guided parameter edits corrected $\mathbf{38\%}$ of IDEA's errors (38/100), compared to $12\%$ for CoT with equivalent natural-language feedback (12/100). This $3.2\times$ gap mirrors the controllability results: natural language cannot reliably translate expert intent into behavioural change, whereas parameter edits carry a mathematical guarantee.

\section{Audit of Spurious and Biased Factors}
\label{app:bias-audit}

A natural concern with LLM-mined factor sets is that they may encode demographic or otherwise problematic signals (e.g., race or gender proxies in credit assessment). Manual inspection of LLM-generated factor sets across 300 sampled instances spanning the five benchmarks revealed zero instances of spurious or biased factors.

Beyond this empirical observation, IDEA offers a \emph{structural} advantage for bias mitigation that prompting lacks: every factor is named, its contribution is quantified by its AME, and any flagged factor can be removed with a mathematically guaranteed zero effect, turning bias auditing from an opaque post-hoc task into a transparent, verifiable one. We note that 300 instances constitute a limited audit; deployment in consequential domains should pair IDEA with domain-specific fairness review.

\section{Computational Overhead Analysis}

\subsection{LLM Query Requirements}

\begin{table}[h]
\centering
\small
\begin{tabular}{lcc}
\toprule
\textbf{Method} & \textbf{Setup (one-time)} & \textbf{Per Instance} \\
\midrule
Vanilla/CoT/Logit & 0 & 1 \\
BIRD & 0 & $N$ \\
IDEA & $\sim$20 + $\min(2^N, 256)$ & $N + T$ \\
\bottomrule
\end{tabular}
\caption{LLM query counts. For IDEA with $N=6, T=50$: setup $\approx$ 84 queries; inference = 56 queries/instance.}
\label{tab:query_counts}
\end{table}
We can also conduct parallel queries and utilize LLM to speed up the process. Furthermore, we would like to emphasize that it is essential to utilize LLM as much as possible to gain a comprehensive understanding of all the knowledge it possesses regarding a query scenario. This is the core and strength of our approach. Therefore, it is imperative to frequently invoke the API of LLM. Efficiency is not the selling point of our method. It's not what we are aiming for. Moreover. if we make decisions based on the trained decision model after the training process is completed, we only need to determine the values of specific conditions. Once determined, the computational cost of our decision-making can be considered almost zero. This is because the computational cost of the logistic regression model is so low that it can be negligible, and it can instantly handle a large number of queries.
IDEA requires more queries than direct methods, but setup costs (factor identification + behavioral probing) amortize across test instances. For 1,000 instances, the per-instance overhead averages to $\sim$56.1 queries.

\subsection{Latency Comparison}

\begin{table}[h]
\centering
\small
\begin{tabular}{lcc}
\toprule
\textbf{Method} & \textbf{Latency (s)} & \textbf{Accuracy (\%)} \\
\midrule
Vanilla & 0.8 & 71.7 \\
CoT & 2.1 & 69.2 \\
IDEA ($T$=50) & 18.4 & 73.2 \\
IDEA ($T$=20) & 8.2 & 72.4 \\
\bottomrule
\end{tabular}
\caption{Per-instance latency with sequential execution (Qwen-3-8B).}
\label{tab:latency}
\end{table}

With parallelized API calls, IDEA ($T$=50) latency reduces to approximately 5 seconds. For latency-sensitive applications, $T=20$ provides a reasonable trade-off, maintaining most accuracy gains at 40\% of the query cost.

\section{Hyperparameter Configuration}
\label{appendix:hyperparams}

Table~\ref{tab:hyperparams} summarizes all hyperparameters used in IDEA.

\begin{table}[h]
\centering
\small
\begin{tabular}{llc}
\toprule
\textbf{Component} & \textbf{Hyperparameter} & \textbf{Value} \\
\midrule
\multirow{3}{*}{EM Algorithm} 
& Precision ratio $\tau_\theta/\tau_\phi$ & 1.0 \\
& Convergence threshold & $10^{-4}$ \\
& Max iterations & 100 \\
\midrule
\multirow{3}{*}{Regularization} 
& $\lambda_1$ (L1 on $\bm{\gamma}$) & 0.01 \\
& $\lambda_2$ (L2 on $\bm{\gamma}$) & 0.001 \\
& $\lambda$ (ranking loss) & 0.1 \\
\midrule
\multirow{2}{*}{Inference} 
& MC samples $T$ & 50 \\
& Sampling temperature & 1.2 \\
\midrule
Probing & Configuration samples & 256 \\
\bottomrule
\end{tabular}
\caption{Hyperparameter configuration for IDEA.}
\label{tab:hyperparams}
\end{table}

\paragraph{EM Parameters.} The precision ratio $\tau_\theta/\tau_\phi = 1.0$ assigns equal trust to model predictions and verbal indications. We searched over $\{0.5, 1.0, 2.0\}$ and found performance stable across this range. Convergence typically occurs within 15--30 iterations.

\paragraph{Regularization.} We select $(\lambda_1, \lambda_2)$ via 5-fold cross-validation on the behavioral probing dataset. The ranking loss weight $\lambda = 0.1$ was chosen from $\{0.01, 0.1, 0.5\}$ to balance regression fit with ordinal consistency.

\paragraph{Inference.} We set $T = 50$ Monte Carlo samples based on diminishing accuracy gains beyond this point. Sampling temperature 1.2 balances diversity against coherence. or each instance, we also make the LLMs run 3 times and record the average scores for each instance in all experiments.

\paragraph{Verbal Mapping Initialization.} Following ~\citep{Budescu2009Improving}, we initialize $\phi$ as: \textit{very unlikely}: 0.05, \textit{unlikely}: 0.15, \textit{somewhat unlikely}: 0.30, \textit{neutral}: 0.50, \textit{somewhat likely}: 0.70, \textit{likely}: 0.85, \textit{very likely}: 0.95.

\section{Sensitivity Analysis}
\label{appendix:sensitivity}

We analyze sensitivity to critical hyperparameters using Qwen-3-8B on COMMON2SENSE.

\begin{table}[h]
\centering
\small
\begin{tabular}{ccc}
\toprule
\textbf{Samples ($T$)} & \textbf{Accuracy (\%)} & \textbf{Std. Error} \\
\midrule
10 & 93.2 & 0.42 \\
25 & 94.3 & 0.28 \\
50 & 94.8 & 0.19 \\
100 & 94.9 & 0.14 \\
\bottomrule
\end{tabular}
\caption{Effect of Monte Carlo sample size.}
\label{tab:mc_sensitivity}
\end{table}

\begin{table}[h]
\centering
\small
\begin{tabular}{ccc}
\toprule
\textbf{$\lambda_1$} & \textbf{Interaction Sparsity (\%)} & \textbf{Accuracy (\%)} \\
\midrule
0.001 & 12.5 & 94.1 \\
0.01 & 45.8 & 94.8 \\
0.1 & 87.5 & 93.6 \\
\bottomrule
\end{tabular}
\caption{Effect of L1 regularization strength.}
\label{tab:l1_sensitivity}
\end{table}

Results indicate IDEA is robust to hyperparameter choices within reasonable ranges. The most sensitive parameter is $\lambda_1$, which controls the interpretability-accuracy tradeoff via interaction sparsity.
\section{Experiment II Ground Truth Construction}
\label{appendix:ground_truth}

\subsection{Dataset Composition}
Our ground truth dataset consists of 500 paired instances. Of these, 350 instances (70\%) are sourced from the existing expert-annotated BIRD dataset ~\citep{DBLP:conf/iclr/0013ZLR25}. The remaining 150 instances (30\%) were newly annotated by experts to ensure adequate coverage of edge cases and complex reasoning scenarios.

\subsection{Annotation Protocol}
We recruited a pool of eight annotators through a university participant pool. All participants were native English speakers holding at least a bachelor's degree. To ensure high-quality data, candidates were screened via a 20-instance qualification task; only those achieving an accuracy of $\geq$85\% were selected (8 of 11 candidates qualified). Selected annotators underwent a 30-minute training session focused on assessing evidential strength before beginning independent annotation.

\paragraph{Instructions and Compensation.}
Annotators were provided with detailed guidelines defining ``evidential strength'' as the logical support a condition provides for a specific outcome. The instructions included definitions for strict dominance and equivalence, accompanied by five anchor examples demonstrating varying degrees of logical support. 
Participants were compensated at a rate of \textbf{\$10} per hour, which was determined to be above the local minimum wage and consistent with fair research standards.

\paragraph{Task Definition.}
For each paired query $(Q_1, Q_2)$, annotators were tasked with selecting one of three labels: (a) $C_1$ provides stronger support, (b) $C_2$ provides stronger support, or (c) Both provide equal support. Each instance was evaluated by three independent annotators, with final labels determined by majority vote.

\subsection{Inter-Annotator Agreement}
We evaluate the reliability of our new annotations using Fleiss' $\kappa$ and percentage agreement metrics, as summarized in Table~\ref{tab:iaa}.

\begin{table}[h]
\centering
\small
\begin{tabular}{lc}
\toprule
\textbf{Metric} & \textbf{Value} \\
\midrule
Fleiss' $\kappa$ & 0.71 \\
Pairwise Agreement & 84.2\% \\
Unanimous Agreement & 67.3\% \\
\bottomrule
\end{tabular}
\caption{Inter-annotator agreement statistics for the 150 newly annotated instances.}
\label{tab:iaa}
\end{table}

The obtained Fleiss' $\kappa$ of 0.71 indicates substantial agreement among annotators ~\citep{landis1977measurement}. Agreement on cases with clear dominance was notably higher (91.4\%), with disagreements primarily concentrated on borderline ``Equal'' instances. To validate consistency with the source data, we performed a re-annotation of 50 randomly sampled BIRD-derived instances, which yielded an agreement rate of 88.0\% (Cohen's $\kappa = 0.79$), confirming robust cross-dataset consistency.

\section{Use of AI Tools}
During the preparation of this work, the authors utilized Large Language Models (LLMs) exclusively for grammatical refinement, proofreading, and stylistic polishing to improve readability. The scientific content, logical reasoning, and conclusions presented in this manuscript remain the sole responsibility of the authors.

\section{Prompting Display}
\begin{figure}[t]
    \centering
    \includegraphics[width=0.5\textwidth]{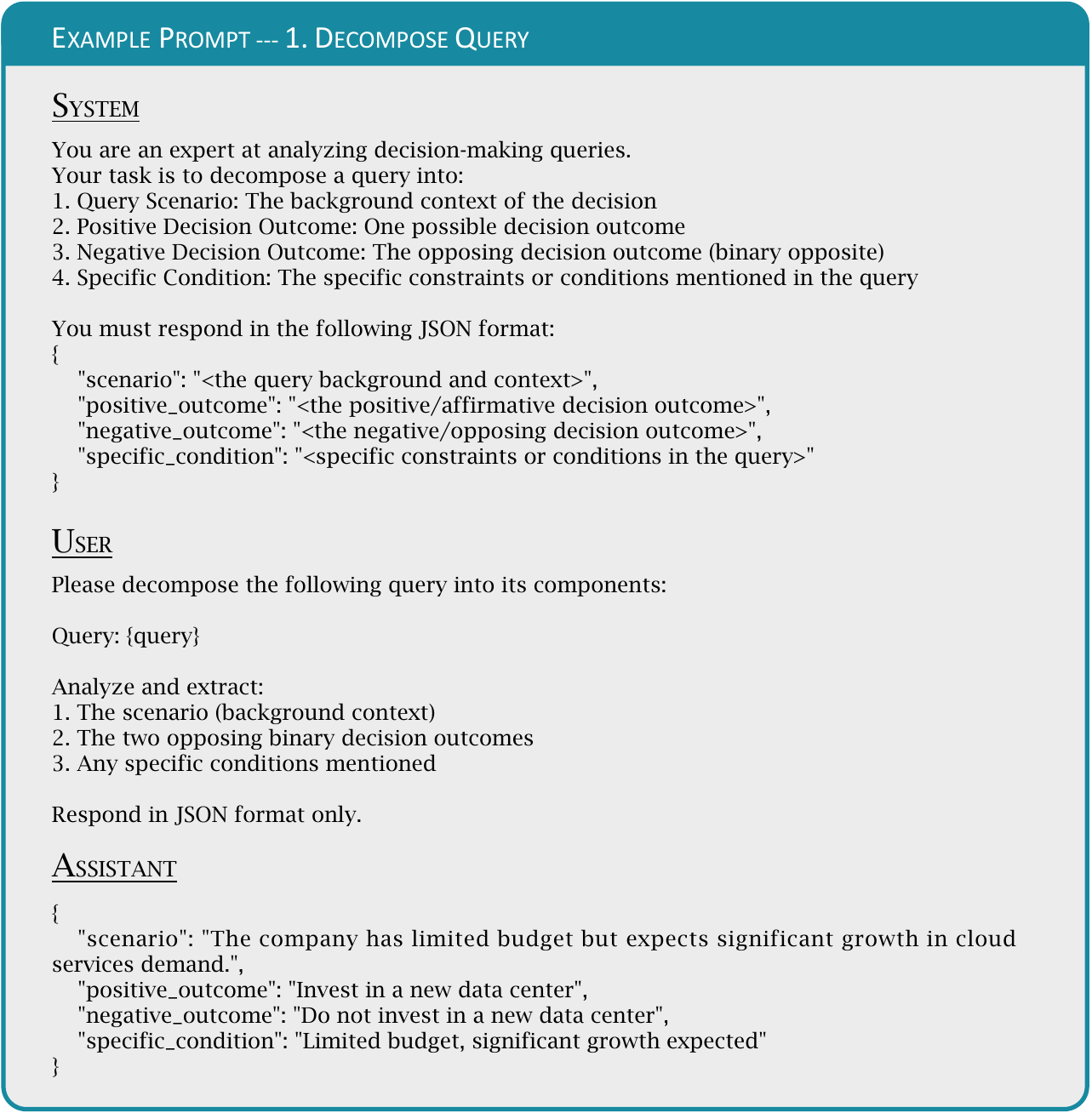}
    \caption{Prompting Example: DECOMPOSE QUERY.}
    \label{fig:prompt1}
\end{figure}
\begin{figure}[t]
    \centering
    \includegraphics[width=0.5\textwidth]{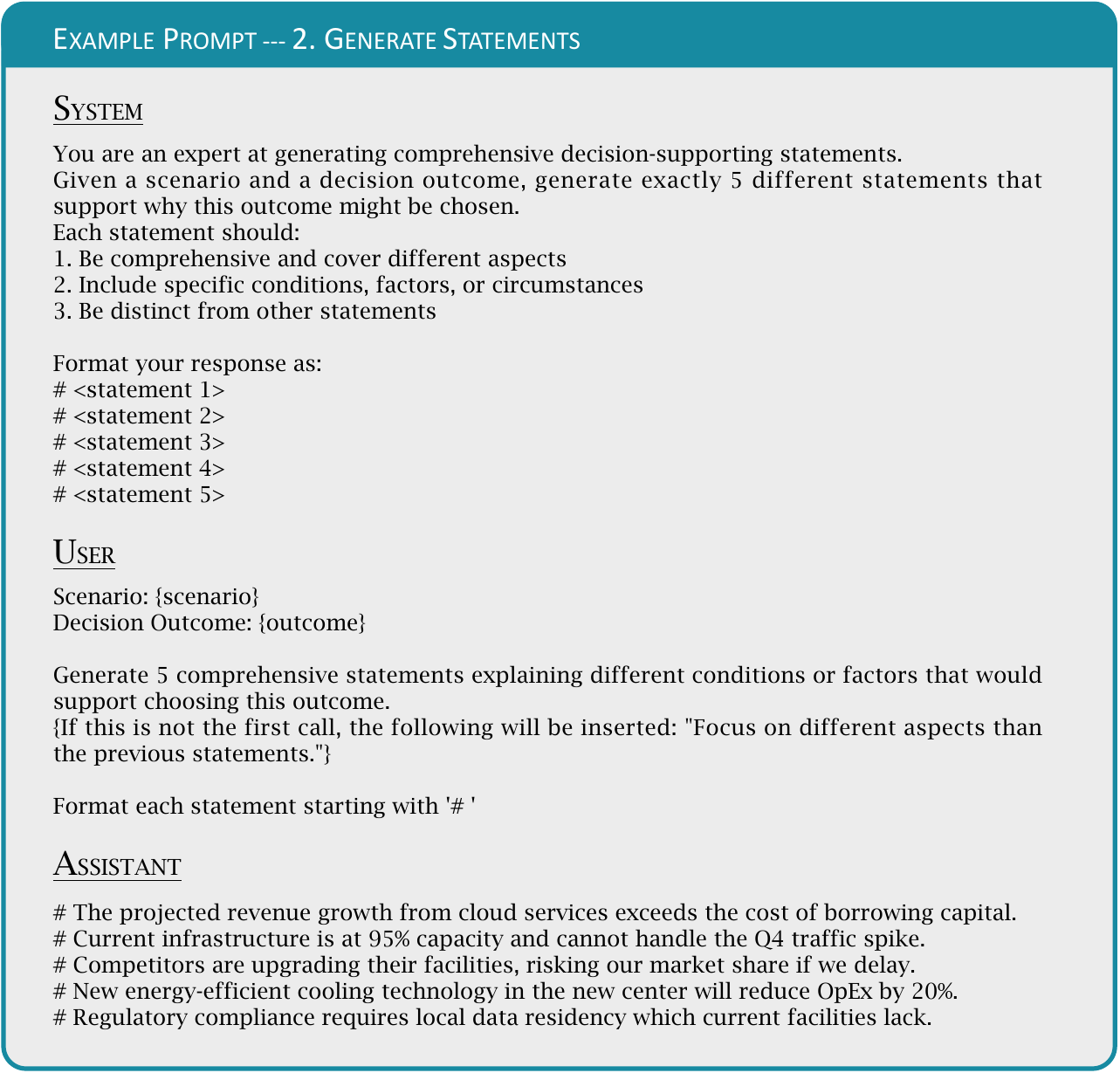}
    \caption{Prompting Example: GENERATE STATEMENTS.}
    \label{fig:prompt2}
\end{figure}
\begin{figure}[t]
    \centering
    \includegraphics[width=0.5\textwidth]{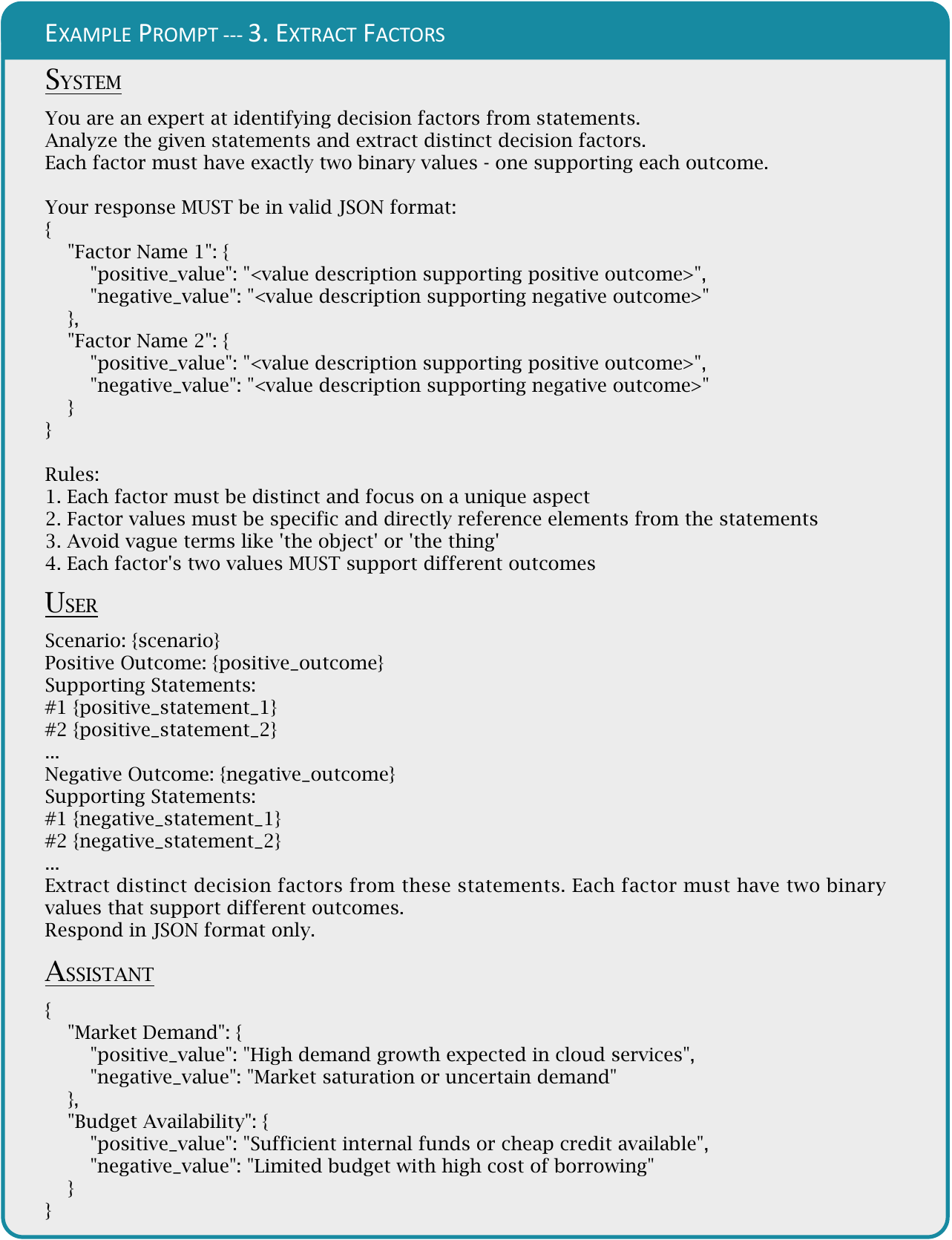}
    \caption{Prompting Example: EXTRACT FACTORS.}
    \label{fig:prompt3}
\end{figure}
\begin{figure}[t]
    \centering
    \includegraphics[width=0.5\textwidth]{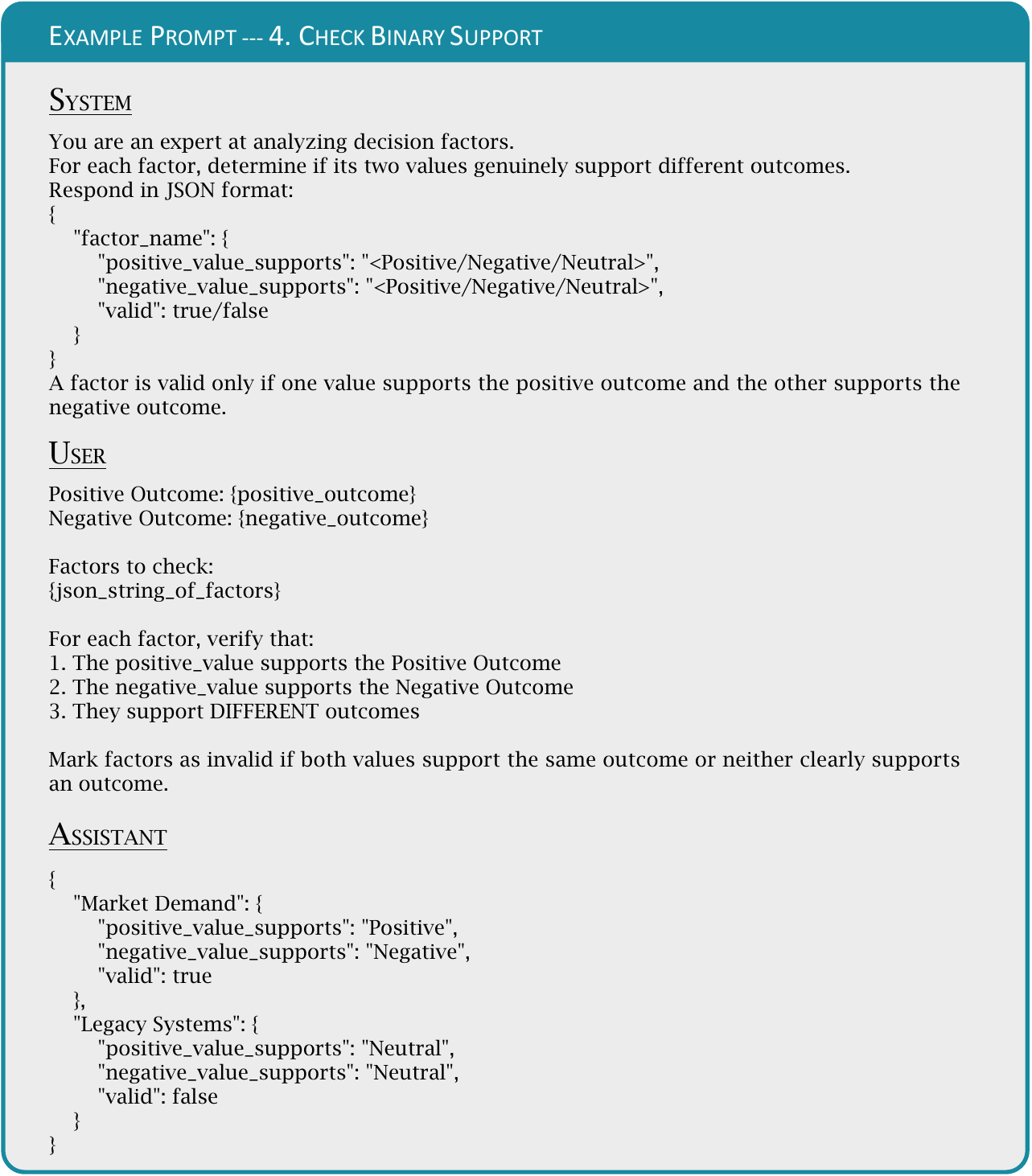}
    \caption{Prompting Example: CHECK BINARY SUPPORT.}
    \label{fig:prompt4}
\end{figure}
\begin{figure}[t]
    \centering
    \includegraphics[width=0.5\textwidth]{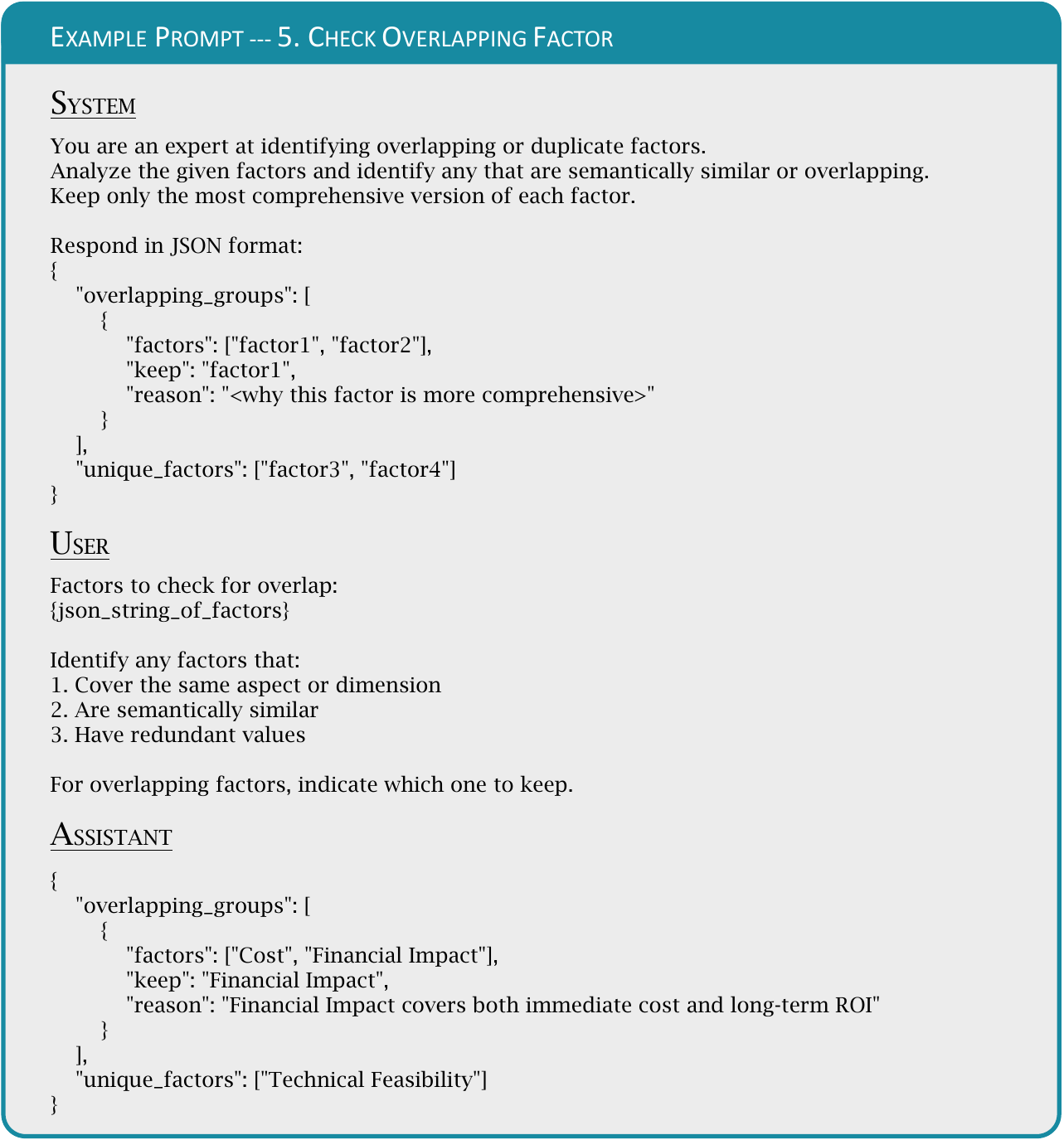}
    \caption{Prompting Example: CHECK OVERLAPPING FACTOR.}
    \label{fig:prompt5}
\end{figure}
\begin{figure}[t]
    \centering
    \includegraphics[width=0.5\textwidth]{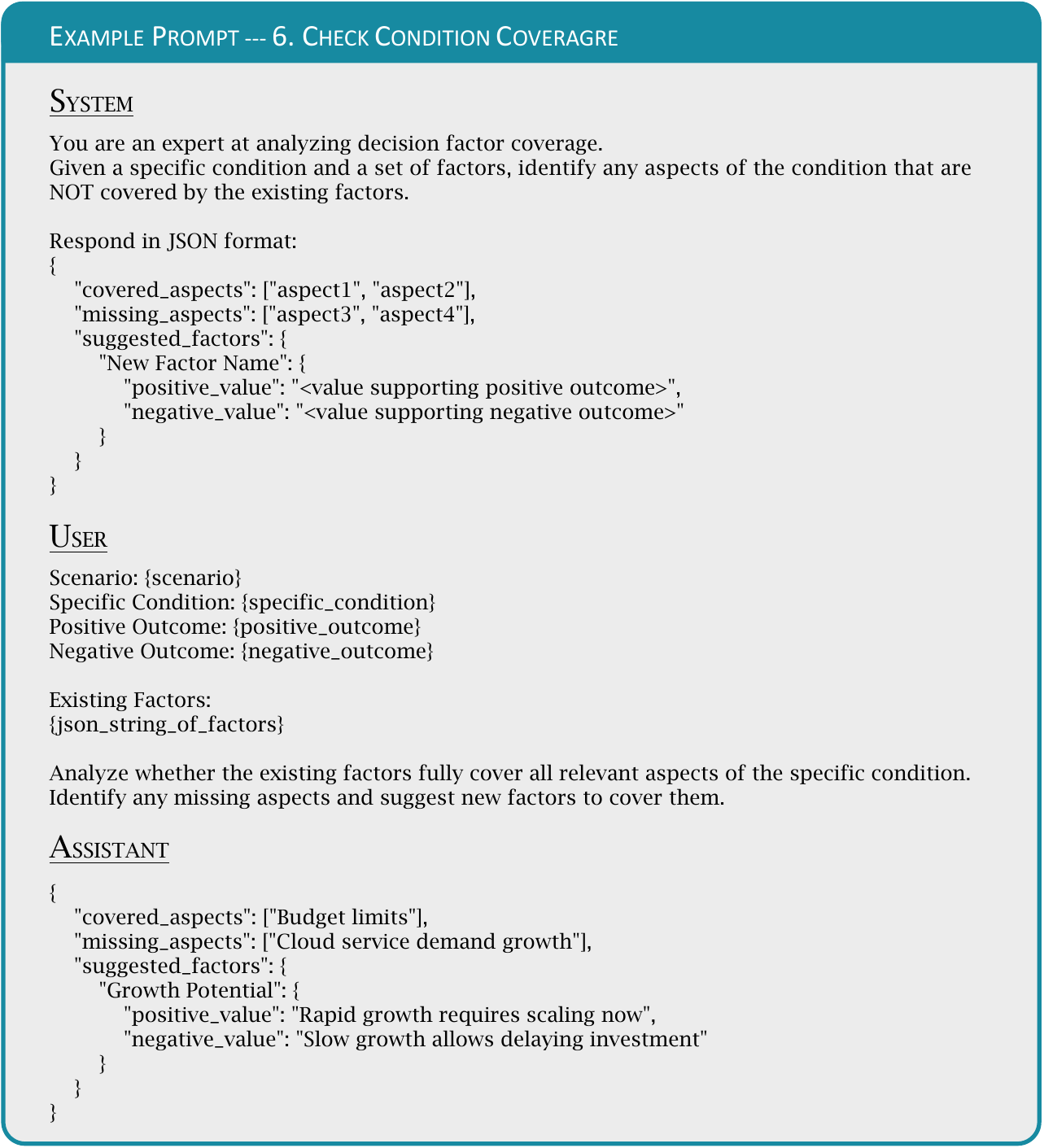}
    \caption{Prompting Example: CHECK CONDITION COVERAGRE.}
    \label{fig:prompt6}
\end{figure}
\begin{figure}[t]
    \centering
    \includegraphics[width=0.5\textwidth]{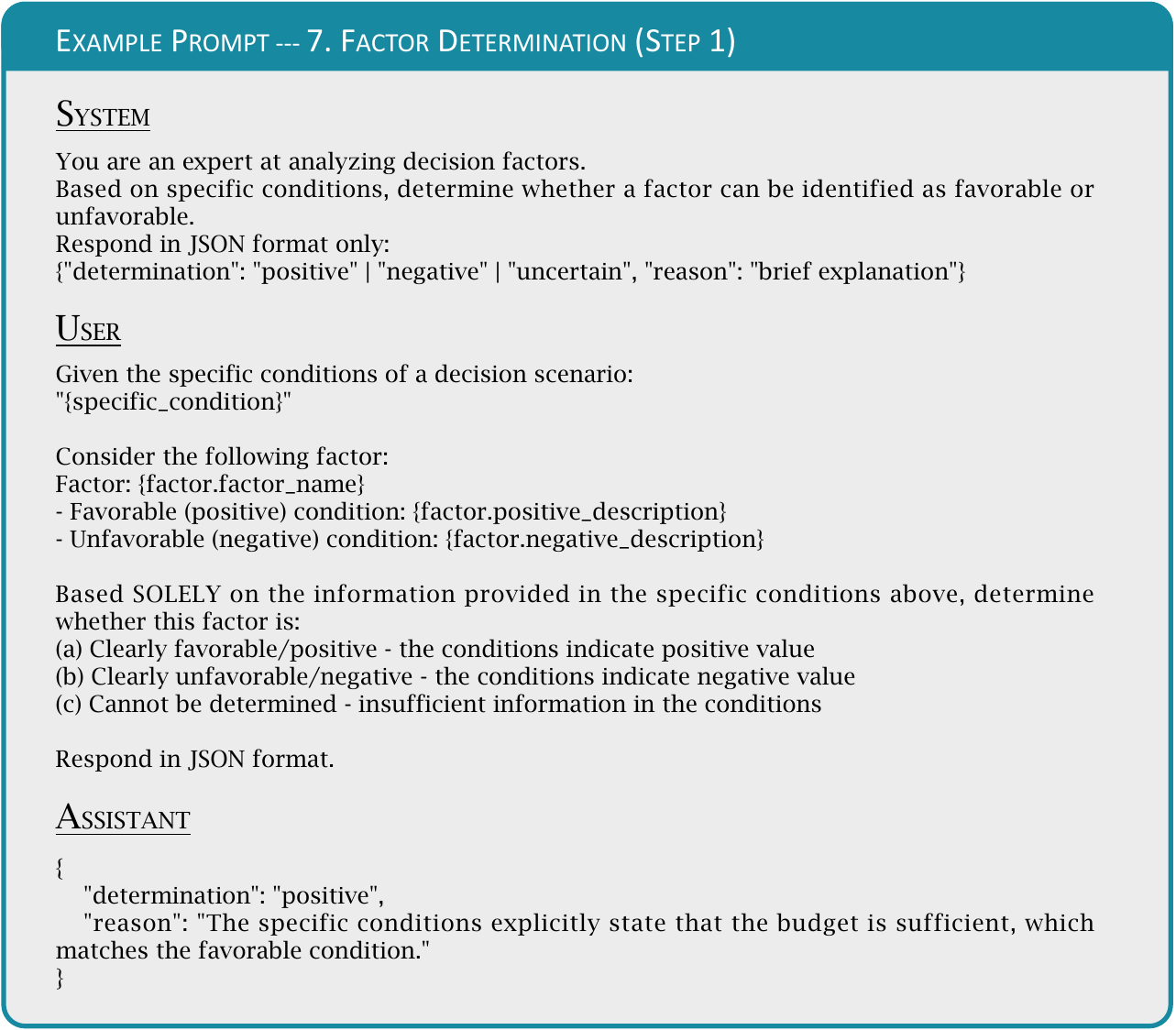}
    \caption{Prompting Example:  FACTOR DETERMINATION (STEP 1).}
    \label{fig:prompt7}
\end{figure}
\begin{figure}[t]
    \centering
    \includegraphics[width=0.5\textwidth]{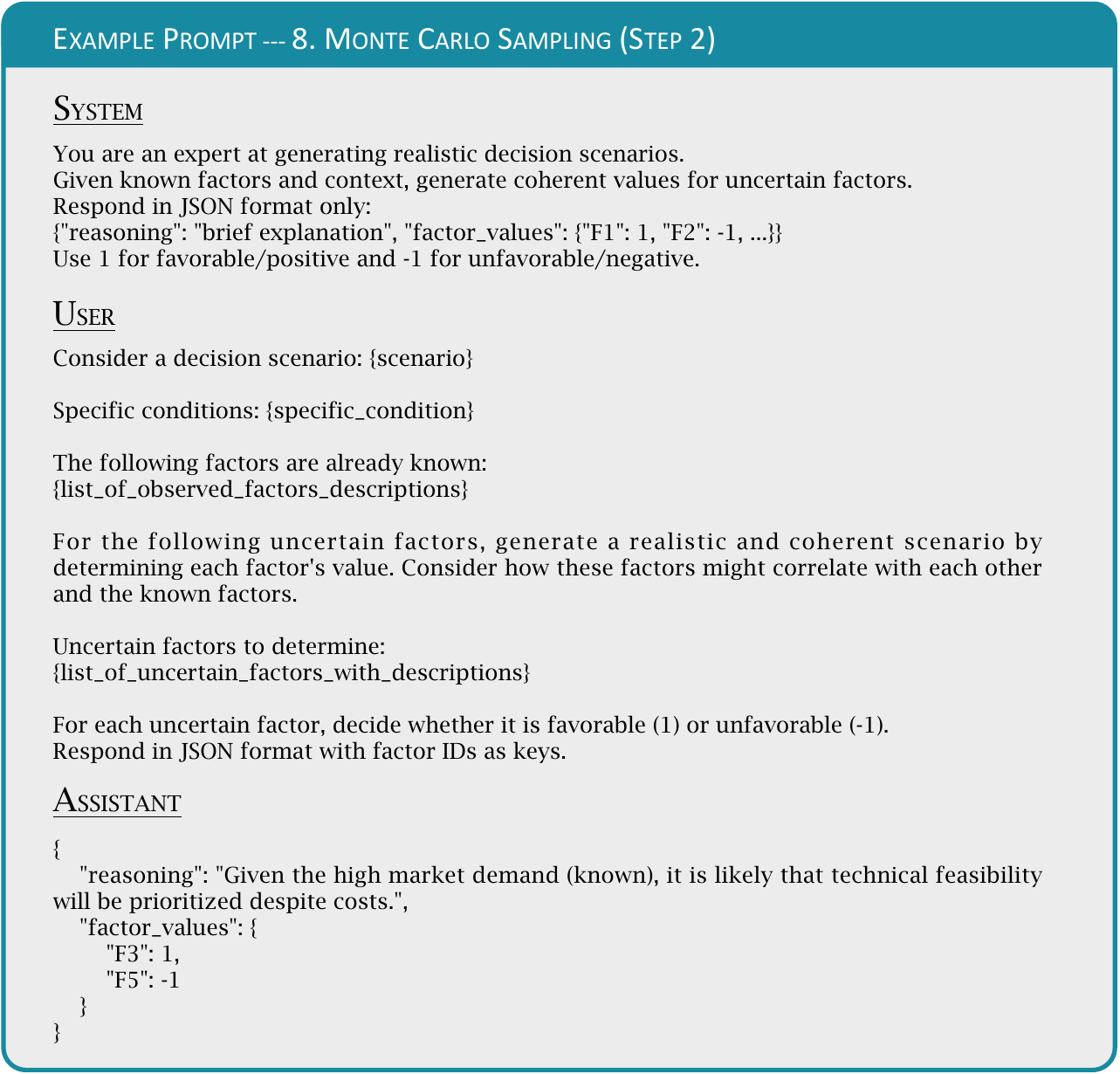}
    \caption{Prompting Example: MONTE CARLO SAMPLING (STEP 2).}
    \label{fig:prompt8}
\end{figure}

\end{document}